\documentclass{article}
\usepackage{ijcai21}

\usepackage{times}

\usepackage{soul}
\usepackage{url}
\usepackage[hidelinks]{hyperref}
\usepackage[utf8]{inputenc}
\usepackage[small]{caption}
\usepackage{graphicx}
\usepackage{amsmath}
\usepackage{booktabs}
\usepackage{comment}
\usepackage{color}
\usepackage{algorithm} 
\usepackage{algpseudocode} 
\usepackage[table,xcdraw]{xcolor}
\usepackage{amsmath,amssymb,amsfonts}
\usepackage{graphicx}
\usepackage{textcomp}
\usepackage{xcolor}
\usepackage{multirow}
\usepackage{colortbl}
\title{Against Membership Inference Attack: Pruning is All You Need}

\author{
Yijue Wang$^1$\and
Chenghong Wang$^2$\and
Zigeng Wang$^1$\and
Shanglin Zhou$^1$\and 
Hang Liu $^3$\and \\
Jinbo Bi $^1$\and 
Caiwen Ding $^1$\and 
Sanguthevar Rajasekaran $^1$\footnote{Corresponding Author}
\affiliations
$^1$University of Connecticut\\
$^2$Duke University\\
$^3$Stevens Institute of Technology \\
\emails
\{yijue.wang, zigeng.wang, shanglin.zhou, jinbo.bi, caiwen.ding, sanguthevar.rajasekaran\}@uconn.edu,
cw374d@duke.edu,\\
hang.liu@stevens.edu
}

\begin{document}

\maketitle

\begin{abstract}

 The large model size, high computational operations, and vulnerability against membership inference attack (MIA) have impeded deep learning or deep neural networks (DNNs) popularity, especially on mobile devices.  To address the challenge, we envision that the weight pruning technique will help DNNs against MIA while reducing model storage and computational operation. In this work, we propose a pruning algorithm, and we show that the proposed algorithm can find a subnetwork that can prevent privacy leakage from MIA and achieves competitive accuracy with the original DNNs. We also verify our theoretical insights with experiments. Our experimental results illustrate that the attack accuracy using model compression is up to 13.6\% and 10\% lower than that of the baseline and Min-Max game, accordingly. 

\end{abstract}

%%
%% The code below is generated by the tool at http://dl.acm.org/ccs.cfm.
%% Please copy and paste the code instead of the example below.
%%

%%
%% Keywords. The author(s) should pick words that accurately describe
%% the work being presented. Separate the keywords with commas.
%\keywords{Deep Neural Network, Membership Inference Attack, weight pruning, Privacy}

%% A "teaser" image appears between the author and affiliation
%% information and the body of the document, and typically spans the
%% page.
% \begin{teaserfigure}
%   \includegraphics[width=\textwidth]{sampleteaser}
%   \caption{Seattle Mariners at Spring Training, 2010.}
%   \Description{Enjoying the baseball game from the third-base
%   seats. Ichiro Suzuki preparing to bat.}
%   \label{fig:teaser}
% \end{teaserfigure}

%%
%% This command processes the author and affiliation and title
%% information and builds the first part of the formatted document.
\maketitle
%\setlength{\textfloatsep}{2pt plus 1.0pt minus 1.0pt}
%\setlength{\floatsep}{2pt plus 1.0pt minus 1.0pt}
%\setlength{\intextsep}{2pt plus 1.0pt minus 1.0pt}

%\vspace{-0.2cm}
\section{Introduction}
Advances in Machine Learning (ML) have
enabled high accuracy in classifications, recommendations, and natural language processing, etc~\cite{he2016deep,vaswani2017attention,9127461}. The success of modern deep neural networks (DNNs) is mainly dependent on the availability of advanced computing power and a large number of data. Machine-Learning-As-A-Service (MLaaS)~\cite{ribeiro2015mlaas} providers such as  Amazon~\cite{kurniawan2018learning}, Microsoft~\cite{gollob2015microsoft}, and Google~\cite{ravulavaru2018google} 
have taken advantage of the aforementioned two availabilities. 
% Recently, to empower users with lower cost and data transmission latency, directly running MLaaS at the network edge has increasing becoming popular. 
By providing black-box interfaces, MLaaS allows individuals or groups to upload data easily, leverage powerful large-scale DNNs, and deploy analytic services via pay-as-you-go or subscriptions using personal computers or mobile devices~\cite{truex2019demystifying}. 
% Recently, to empower users with  lower cost and data transmission latency, directly running MLaaS at the network edge has increasing becoming popular.
% empowers users with the fastest speeds, lowest latency, and highest processing power possible, optimizing machine learning. 
% By providing black-box interfaces, MLaaS allows individual users to upload their data, build models and deploy analytic services~\cite{truex2019demystifying}. 
% Given this landscape, the availability of such black-box machine learning service poses new challenges in different domains, which raises great interests in both academia and industry.

However, there are two main challenges. (i) MLaaS raises privacy concerns on sensitive data such as patient treatment records. Even though the DNN model structures are in black-box, MLaaS can leak sensitive information about training data used to build back-end models.
% that the use of sensitive data in providing MLaaS is inevitable, such as patient treatment records, which are extremely sensitive. 
% Thus, MLaaS raise certain safety and privacy concerns even though the ML model structures are in black-box. 
% If the fundamental design of such systems ignores the adversarial threats, even with black-box access, such service can leak sensitive information about the training data used to build back-end models\cite{shokri2017membership}. 
For instance, membership inference attack (MIA) ~\cite{shokri2017membership} is one of the critical inference attacks in exploiting the aforementioned vulnerability. By using MIA, the adversary monitors the distinctive behavior of back-end models by repeating sophisticated designed inference requests to further exploit information about the training data.  (ii) DNN models are evolving fast in order to satisfy the diverse characteristics of broad applications. As the layers of DNNs get deeper and the model size of DNNs gets larger, the large computational operations and model size introduce substantial data movements, limiting their ability to provide a user-friendly experience on mobile devices~\cite{krizhevsky2012imagenet,hinton2012deep}. 

% we need to determine whethe the data 
%\textcolor{red}{(add a short description of why DP does not work for MIA?)}
%\cw{I suggest breif introduce the two defense MinMax and DP, then point out their limitations}
There are several mechanisms have been developed to address the MIA challenges.
% One major defense mechanism against general Inference attack is 
Differential privacy (DP), a major privacy-preserving mechanism against general Inference attack, which is based on adding noises into gradients or objective function of the training model, has been applied in different machine learning models ~\cite{abadi2016deep,zhang2019predictive,rahman2018membership}. Although the robustness of DP has been proven,
% d against inference attacks, 
the utility cost (e.g., creating indistinguishable non-membership datasets, calculating the bound for the function sensitivity) of DP is hard to be limited as acceptable since it imposes a significant accuracy loss for protecting complicated models as well as on high dimensional data when noise is considerable. 
 
Another defense mechanism is game theory, e.g., Min-Max game~\cite{nasr2018machine}, which guarantees the information privacy. 
The maximum gain of inference model is considered as a new regularization called {\it adversarial regularization} and will be minimized with the training model loss. Unfortunately, the Min-Max game introduces extra computational operations in addition to the classifier training process. 

Finally, yet importantly, neither DP nor Min-Max game addresses the second challenge, i.e., large computational operations and model size in DNNs.
% , which is proposed by Cynthia Dwork on 
% \cite{dwork2006calibrating, dwork2014algorithmic}. It has been applied in different machine learning models and deep learning models\cite{abadi2016deep, bassily2014private,zhang2019predictive, chaudhuri2011differentially, rahman2018membership}.
% Although the robustness has been proven,
% % d against inference attacks, 
% DP mechanism’s cost of utility is hard to be limited as acceptable. the existing mechanisms would impose a significant accuracy loss for protecting complicated models as well as on high dimensional data when noise is large. Another mechanism against the MIA is game theory which is used to guarantee the information privacy and the state of art of it is min-max game. In such a framework, the maximum gain of inference model is considered as a factor and will be minimized with training model loss. However, game based method (i.e min-max game) typically introduces extra computational operations in addition to the training approaches. Moreover, L2-Regularizer\cite{shokri2017membership}, dropout\cite{salem2018ml}\cite{srivastava2014dropout}, model stacking\cite{salem2018ml}, Memguard\cite{jia2019memguard} are other technique proposed against MIA. The limitation of these method are also based on the formal utility loss guarantee and the computation performance improving. 

In this paper, to simultaneously address the two challenges,
% limitations of previous MIA defense methods, 
we develop our pruning algorithm that is optimized for the dual objectives of privacy and efficiency by finding a subnetwork from a sufficiently over-parameterized random network.
We present the main contributions of our work as follows:
% therefore againt the MIA. In this work, we thorougly inveåçtigate
\leftmargini=4mm
% \leftmarginii=6mm
\begin{itemize}
    \item To the best of our knowledge, this work is the first attempt to simultaneously address the challenges of large model size,  high computational cost,  and vulnerability against  MIA on DNNs. We jointly formulate weight pruning and MIA as MIA-Pruning and provide an analytic solution strategy for the problem.
       
    \item We show that our pruning algorithm can find a subnetwork that can prevent the privacy leakage from MIA and achieves competitive accuracy with the original DNNs.
    
    \item %We investigate the MIA-Pruning to evaluate if weight pruning 
    % has the same effectiveness
    %can also reduce attack accuracy as the Min-Max game.
    % , i.e., reduce attack accuracy. 
    We show that our pruning algorithm performs better than baseline (without defense and pruning) and Min-Max game, i.e., further reduces attack accuracy. We also investigate the combination of weight pruning and Min-Max game and show that the combination will further enhance DNN model privacy.

\end{itemize}

Experimental results show that our MIA-Pruning can help against MIA while simultaneously achieving model storage and computational complexity reduction within a very small accuracy loss.
Our proposed method significantly outperforms DP on MIA. 
Since weight pruning reduces the number of parameters, the proposed MIA-Pruning enables faster DNN computation than prior works.

\section{Related Work }

\subsection{Weight Pruning} State-of-the-art DNNs contain multiple cascaded layers and at least millions of parameters (i.e., weights) for the entire model ~\cite{he2016deep,vaswani2017attention}.
Prior works have focused on developing DNN {weight pruning} algorithms such as weight pruning ~\cite{zhang2018systematic,frankle2018lottery,zhou2019deconstructing,ramanujan2020s} (i.e., removing weights with specific dimensions or with any desired weight matrix shapes) utilizing different regularization techniques to explore sparsity. Blalock~\cite{blalock2020state} introduced ShrinkBench for standardized evaluations of pruning methods.

\subsection{Defense Mechanism against MIA}
% We summarize multiple MIA defense mechanisms as following.
One defense direction is using game theory to protect privacy~\cite{nasr2018machine,alvim2017information,shokri2015privacy,shokri2012protecting}. Most of the game theory-based mechanisms minimize the privacy loss against the most potent attacker by converting the utility function into the Min-Max optimization problem. 
%After Generative Adversarial Network (GAN) being proposed by~\cite{goodfellow2014generative}, some new algorithms for solving min-max problem while training DNN model. For instance, using a similar framework as GAN,

\cite{nasr2018machine} proposed a Min-Max game mechanism and formulated the gain of MIA as a new regularization, which is maximized while the classifier's loss is minimized. We use it as a comparison with our experimental results. 

{DP is another major defense mechanism against MIA.} There are multiple DP-based defense mechanisms~\cite{chaudhuri2011differentially,abadi2016deep,iyengar2019towards}, by adding noises into gradients or the objective function of the training model. {However, the existing mechanisms would impose a significant accuracy loss for protecting complicated models as well as on high dimensional data when the noise parameter $\epsilon$ is large.} 
%Differential privacy mechanisms are difficult to achieve with negligible utility loss, where utility loss is related to creating the same distribution's states of all input data, and also computing the gradient noise with a narrow bound. 

There are some other defense directions. For example, the model stacking~\cite{salem2018ml} mechanism made a combination of multiple classifier results to prevent the attacker from inferring a single target classifier.  MemGuard mechanism~\cite{jia2019memguard} randomly added noise on the target classifier prediction. 

The existing defenses have at least one of the limitations: 1) they have typical extra computations, such as extra weight storage and noise calculations. That means these mechanisms introduce extra computational operations in addition to the training approaches. 2) they achieve privacy protection with {significant utility loss.}

\begin{figure}[t]
\centering
	\includegraphics[width=0.45\textwidth]{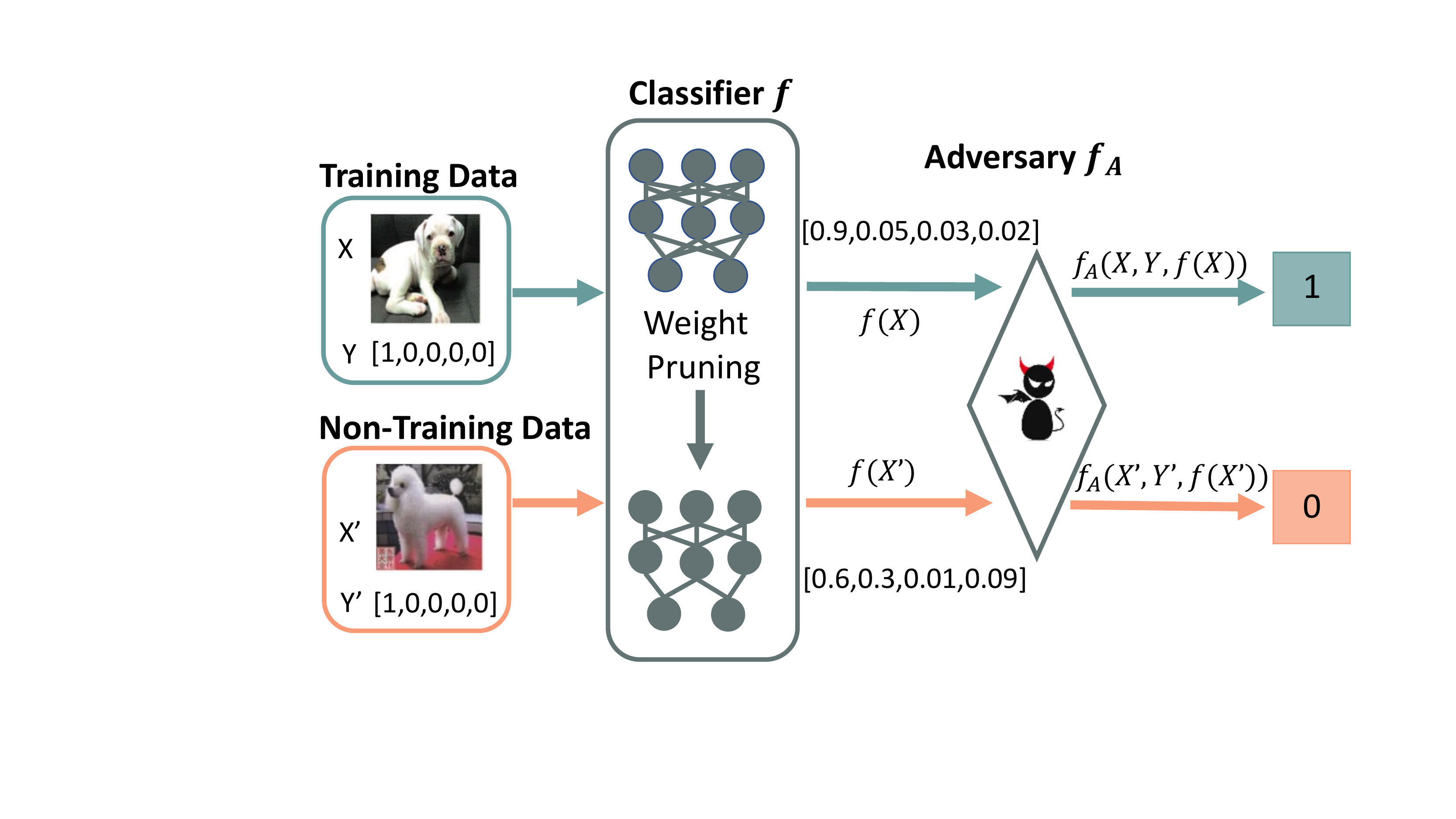}

    \caption{{ Illustrative diagram of using weight pruning against MIA.} }
    \label{fig:system}
\end{figure}

\section{MIA-Pruning: Problem Statement}
In this work, we investigate the following question: {\em Will an effective DNN weight pruning technique help against MIA while simultaneously achieving model storage and computational complexity reduction?
% within very small accuracy loss?
} 
%We start with formulating the joint problem of weight pruning and MIA: (i) MIA-Pruning against MIA; (ii) MIA-Pruning combined with Min-Max game against MIA.
\subsection{Problem Formulation}
%\noindent\textbf{MIA.}
\subsubsection{MIA}
For the target machine learning model, we consider the classification model in this work. Let $f$ denotes the target classification model, $x$ denotes a data point, and $f(x)$ denotes the output of $f$ on data $x$. $f(x)$ is a one-hot vector of probabilities of $x$ belonging to $k$ classes. We consider the MIA problems in a black-box condition, which means the adversary can not access the classification model's parameters but can only observe the input and output of the classification model. We assume that the adversary has access to some data records from the training set and the predictions from the black-box DNN target model. Based on the difference between the model's prediction on the training dataset and the non-training dataset, the adversary can determine whether a data record belongs to the model's training dataset or not.  Figure~\ref{fig:system} shows an illustrative diagram of using weight pruning against MIA in DNNs. We use $f_A$ to denote the adversarial inference model $f_A: x\times y \times f(x) \longrightarrow [0,1]$. $f_A$ takes the feature of the data $x$, the label of the data $y$, and the prediction of classification model $f(x)$ as inputs. And $f_A$ outputs the probability of data $(x,y)$ belonging to the training set $D$ or the non-training set $D^\prime$. The probability distributions of samples in $D$ and $D^\prime$ are $P_D$ and $P_{D^\prime}$, respectively. The gain function of the inference model $f_A$ given the classification model $f$ can be written as:
\begin{equation}
    \begin{aligned}
        G_{f}(f_A)=& \underset{(x,y)\sim P_D}{\mathbb{E}}\big[\log(f_A(x,y,f(x)))\big]\\
        &+\underset{(x,y)\sim p_{D^\prime}} {\mathbb{E}}\big[\log(1-f_A(x,y,f(x)))\big]
    \end{aligned}
\label{eq:gain}
\end{equation}
The first expectation computes the inference model's accuracy in predicting training data (members), and the second expectation computes the accuracy of the inference model on predicting non-training data (non-members). The underline probability $P_D$ and $P_{D'}$ is normally not known. The empirical gain can be calculated by simply sampling data from the training set and validation set.

\subsubsection{MIA-Pruning against MIA}
The objective is to build a defense system against MIA by pruning the model weights so that the model predictions for the training dataset and non-training dataset are not distinguishable. In this case, it becomes more difficult for the adversary to determine where an observed data record belongs to. Finally, the risk of membership privacy loss is reduced. Ideally, the adversary can only make a determination by random guess. At the same time, the classification accuracy of the model will not be or slightly be affected. In other words, the utility cost of defense (e.g., classification accuracy loss) is negligible.

Formally, the problem can be formulated as:
\begin{equation}
\begin{aligned}
    \underset{\{{\bf{W}}_{i}\},\{{\bf{b}}_{i}\}}{\text{argmin}} \quad
    &\mathcal{L} (f(\{{\bf{W}}_i\},\{{\bf{b}}_i\};x),y)\\
     \text{s.t.} \quad &\mathbf{W}_i \in \{\mathbf{W}_i | card(\mathbf{W}_i) \leq n_i\}\\
     &\{n_i\} = \underset{\{n_i\}}{\text{argmin}} \quad \underset{f_A}{\text{max}} G_{f(\{{\bf{W}}_i\},\{{\bf{b}}_i\})}(f_A)
\end{aligned}
\label{eq:mcmia}
\end{equation}
where $\{\mathbf{W}_i\}$ and $\{\mathbf{b}_i\}$ are the weights and biases of each layer, and $\mathcal{L} (f(\{{\bf{W}}_i\},\{{\bf{b}}_i\};x),y)$ is the loss of the classification model. $card(\mathbf{W}_i)$ is the cardinality of weights in each layer, which returns the number of non-zero weights. $n_i$ is the desired number of non-zero weights for each layer, which regularizes the strength of compression.   

\subsubsection{MIA-Pruning \& Min-Max Game against MIA}
For the further step, we consider the combination of weight pruning and Min-Max game ~\cite{nasr2018machine} to strengthen the defense efficiency against MIA, and the optimization objective would become:
\begin{equation}
\begin{aligned}
    \underset{\{{\bf{W}}_{i}\},\{{\bf{b}}_{i}\}}{\text{argmin}} \quad
    &\mathcal{L} (f(\{{\bf{W}}_i\},\{{\bf{b}}_i\};x),y)\\
    &+ \gamma \text{ } \underset{f_A} {\text{max }} G_f (f_A(f(\{{\bf{W}}_{i}\},\{{\bf{b}}_{i}\};x),y))  \\
     \text{s.t.} \quad &\mathbf{W}_i \in \{\mathbf{W}_i | card(\mathbf{W}_i) \leq n_i\}\\
     &\{n_i\} = \underset{\{n_i\}}{\text{argmin}} \quad \underset{f_A}{\text{max}} G_{f(\{{\bf{W}}_i\},\{{\bf{b}}_i\})}(f_A)
\end{aligned}
\label{eq:mcmia+minmax}
\end{equation}
where $\gamma$ is the weight of the maximum gain of the inference model. The difference between Equation \ref{eq:mcmia} and \ref{eq:mcmia+minmax} is that the loss in \ref{eq:mcmia+minmax} when updating the classification model also includes the maximum possible inference gain given the current classification model. The problem of \ref{eq:mcmia} and \ref{eq:mcmia+minmax} can be written as Equation 14 in Appendix Section 2.

%\vspace{-0.2cm}
\section{MIA-Pruning: Methodology}

\subsection{Solution Strategy}
To find the optimal compression ratio or $n_i$, we set different $n_i$ in a range and choose $n_i$ that can achieve the minimum $G_f(f_A^*)$. For the Min-Max game, the minimization and maximization problems need to be solved jointly to reach an equilibrium point. As shown in algorithm \ref{algo}, for a fixed classification model $f$, we use a subset of training data and non-training data to train the inference model $f_A$ to find the best attack to $f$. Then for the current $f_A$, we minimize the loss in Equation \ref{eq:mcmia} and \ref{eq:mcmia+minmax} to find the most defensive model $f$ against $f_A$. Note that for MIA-Pruning, the for-loop to update $f_A$ is not needed. For an arbitrary $f_A$, it is easy to optimize it empirically by Stochastic Gradient Descent (SGD). However, it is difficult to solve the optimization problem of $f$. We modified the Alternating Direction Method of Multipliers(ADMM)~\cite{zhang2018systematic} to solve it, of which the details can be found in Appendix Section 3.

\subsection{Theoretical Analysis}
\subsubsection{Pruning Convergence Analysis }
We suppose the target network $f(x)$ as:
\begin{equation}
    f(x)=\mathbf{W}_{n}^f\sigma (\mathbf{W}_{n-1}^f...(\sigma(\mathbf{W}_1^f x))
    \label{def_f}
\end{equation}
and we define the original network $g(x)$ as:
\begin{equation}
    g(x)= \mathbf{W}_{2n}^g\sigma(\mathbf{W}_{2n-1}^g...\sigma(\mathbf{W}_1^g(x))
    \label{def_g}
\end{equation}
where $\mathbf{W}_i^f$,$\mathbf{W}_j^g$ is the randomized weight matrix at $i$-th layer of $f$ and $j$-th layer of $g(x)$. And $\sigma(\cdot)$ is the activation function.\\
A pruned network $\hat{g}(x)$ can be presented as :
\begin{equation}
    \hat{g}(x) = (\mathbf{P}_{2n} \odot \mathbf{W}_{2n}^g)\sigma(\mathbf{P}_{2n-1} \odot \mathbf{W}_{2n-1}^g)...\sigma(\mathbf{P}_{1} \odot \mathbf{W}_{1}^gx)
\end{equation}
where $\mathbf{P}_i$ is the pruning matrix in $i$-th layer.

{\bf Theorem 1.} {\it For every network $f$ defined in Eq. \ref{def_f} with depth l and $\forall i \in \{1,2,\dots,n\}$. Consider g defined in Eq.\ref{def_g} is a randomly initialized neural network with 2n layers, and width $poly(d,n,m,1/\epsilon,log1/\delta)$, where $d$ is input size, n is number of layers in $f$, m is the maximum number of neurons in a layer.  The weight initialization distribution belongs to uniform distribution in range [-1,1]. Then with probability at least $1-\delta$  there is a weight-pruned subnetwork $\hat{g}$ of g such that:} 
\begin{equation}
 \underset{x \in \chi, \left \| W \right \| \leq 1}{\text{sup}} {\left \| f(x) - \hat{g}(x) \right \|} \leq \epsilon   
\end{equation}

The full proof of Theorem 1 is in Appendix Section 1. Using Theorem 1, we know that for every bounded distribution and every target network with bounded weights, there is a subnetwork with the competitive accuracy of the original sufficiently over-parameterized neural networks.

\subsubsection{Inference Model's Gain Function Analysis }
According to \cite{nasr2018machine}, we rewrite the gain function of the inference model in the form of probability distribution:
%\vspace{-2mm}
\begin{equation}
    \begin{aligned}
        &G_{f}(f_A)=\int _{x,y}[P_{D}(x,y)p_{f}(f(x))\log(f_A(x,y,f(x))) \\
        &+P_{ D'}({x},{y})p'_f(f(x))\log(1-f_A(x,y,f(x))]dxdy
    \end{aligned}
\label{eq:gain_p2}
\end{equation}
where $D$ is the training set and $D'$ is the non-training set. $p_f$ and $p_f'$ are the probability distribution of the classification model $f$'s output for training data and non-training data.

For a given classification model $f$ and data sampled from a known probability distribution, the optimal determination solution for the inference model $f_A$ is \cite{goodfellow2014generative,nasr2018machine}: 
\begin{equation}
    f_A^*(x,y,f(x))= \frac{p_f(f(x))}{p_f(f(x))+p'_f(f(x^\prime))}
\end{equation}
Therefore, by substituting $f_A^*$ in the Equation \ref{eq:gain}, the gain function of $f_A^*$ can be written as:
\begin{equation}
    \begin{aligned}
        G_{f}(f_A^*)&= \underset{(x,y)\sim P_D}{\mathbb{E}}\big[\log(\frac{p_f(f(x))}{p_f(f(x))+p'_f(f(x))})\big]\\
        &\quad +\underset{(x,y)\sim p_{D^\prime}} {\mathbb{E}}\big[\log(1-\frac{p_f(f(x))}{p_f(f(x))+p'_f(f(x))})\big]\\
        %&=\underset{(x,y)\sim P_D}{\mathbb{E}}\big[\log(\frac{p_f(f(x))}{p_f(f(x))+p'_f(f(x))})\big]\\
        %&\quad +\underset{(x,y)\sim p_{D^\prime}} {\mathbb{E}}\big[\log(\frac{p'_f(f(x))}{p_f(f(x))+p'_f(f(x))})\big]\\
        &=-\text{log}(4) + 2\cdot JS(p_f(f(x)) || p'_f(f(x)))
    \end{aligned}
\end{equation}
Where $JS(p_f(f(x)) || p'_f(f(x)))$ is the Jensen–Shannon divergence between the two distributions. Since $JS(p_f(f(x)) || p'_f(f(x)))$ is always non-negative and equals 0 if and only if $p_f(f({x}))=p'_f(f({x'}))$, the global minimum value that $G_{f}(f_A^*)$ can possibly have is -log(4) if and only if 
$p_f(f({x}))=p'_f(f({x'}))$ ~\cite{goodfellow2014generative}. This means that the prediction of classification model $f$ for both the training set and non-training set has the same probability distribution. In this case, the classification model can be totally protected from MIA, the inference model can only flip a coin to make the determination with the possibility of 0.5.
We use $d$ to represent the Jensen–Shannon divergence $JS(p_f(f(x)) || p'_f(f(x)))$ between the probability distributions of $f$'s outputs for the training set and non-training set. The larger $d$ is, the higher the maximum gain of the reference model is. In other words, the more vulnerable the classification model is. Thus, any methods that can smaller $d$ can have a defending effect against MIA. Intuitively, weight pruning can prevent over-fitting. Thus it will have a smaller $d$.\\

\begin{algorithm}[t]
\small
	\caption{The Process of MIA-Pruning} 
	\begin{algorithmic}[1]
	\For {$\{n_i\}$ in $(\{n_{min}\}, \{n_{max}\})$}
		\For {{\it epoch} in {\it epochs}}
			\For {$t$ in $iterations$}
				
				\State Get a random mini-batch $S \subset D$.
				\State Get a random mini-batch $S' \subset D'$.
				\State Update $f_A$ to minimize $-G_f(f_A)$ using SDG. 
		    \EndFor
			\State Get a random mini-batch $S'' \subset D$, $S''\neq S$.
			\State Update $\{{\bf{W}}_i^t\}, \{{\bf{b}}_i^t\}$  to minimize Loss in Equation \ref{eq:mcmia} 
			\State Prune weight by update $\{{\bf{W}}_i^t\}$ to $\{{ P_i^t\odot\bf{W}}_i^t\}$
		\EndFor
	\State OUTPUT $(\{\mathbf{W}_i\},\{ \mathbf{b}_i\}, G_f(f_A))$.
	\EndFor
	\State OUTPUT $(\{\mathbf{W}_i^*\}, \{\mathbf{b}_i^*\}, G_f^*(f_A)) = min(G_f(f_A))$.
	\end{algorithmic} 
	\label{algo}
\end{algorithm}

%%%%%%%%%%%%%%%

%For MIA-Pruning, weight pruning makes the parameter space of the classification model sparse so that it reduces the risk of over-fitting the model to some features specific to the training set. Intuitively, weight pruning can make the model more generalized, thus the predictions on the training set and non-training set more similar, and the divergence of $p_f(f({x}))$ and $p'_f(f({x}))$ smaller.\\
%For MIA-Pruning + Min-Max game, it further strengthens the defense against MIA by adding the maximum gain to the minimization objective. For each iteration, the maximization finds the strongest inference model for the current classification model,  and the minimization finds the most defensive classification model against the current inference model. The parameter $\gamma$ controls the relative importance of classification accuracy and MIA defense. It has been shown that applying the Min-Max game during machine learning training can guarantee membership privacy \cite{goodfellow2014generative}.\\

%To summarize, MIA-Pruning can constrain MIA's gain, which makes it difficult for the adversary to distinguish the training data record and non-training record based on the data and the classification model's output. Meanwhile, MIA-Pruning minimizes the utility loss (the decrease of classification accuracy) and reduces computation complexity and data storage. The algorithm is showed at Algorithm \ref{algo}

%\vspace{-0.2cm}
\section{Evaluation}
%\vspace{-0.1cm}
In this section, we apply MIA-Pruning to different classification models with various DNN structures, mainly from two perspectives: the defense performance of our model and the computation cost benefit we obtain.
%Specifically, we address the following problems:
%\begin{itemize}
%    \item Could MIA-Pruning reduce attack accuracy with a negligible utility loss?
%    \item Does MIA-Pruning provide a substantial computational performance improvement over unoptimized approaches?
%\end{itemize}

% \subsection{Datasets}
%We use four different machine learning benchmark datasets:
%MNIST, CIFAR-10, CIFAR-100, ImageNet.
% MNIST is a handwritten digits database, which includes 60,000 training images and 10,000 testing images in 10 classes. Each record is composed of 28 $\times$ 28 gray-scale pixels.
% CIFAR-10 consists of 50,000 training images and 10,000 test images in 10 classes. Each image is composed of 32 $\times$ 32 color pixels. CIFAR-100 is the same but has 100 classes.
% ILSVRC2012 is a subset of the large hand-labeled ImageNet dataset, including 1.2 million images in the training set and 50,000 images in the testing set in 1,000 classes. 

\subsection{Classification Model}

%--------------

To evaluate our proposed method, we apply MIA-Pruning on different DNN models, including LeNet-5, VGG16, MobileNetV2, ResNet-18, on different datasets (e.g., MNIST, CIFAR-10, CIFAR-100, ImageNet). 
% \subsection{Classification Models}
We use {LeNet-5} on MNIST dataset, and {VGG-16}, {MobileNetV2} and {ResNet-18} to classify CIFAR-10 and CIFAR-100 dataset.  We also use {MobileNetV2} and {ResNet-18} models on the ImageNet dataset to show the scalability of our proposed method. 

We set the plain training of the classification model, i.e., without adversary training or MIA-Pruning, as our baseline. We compare the classification accuracy of the classification (target) model and the attack accuracy of the inference (adversary) model between MIA-Pruning, Min-Max Game, and baseline. 
%The classification accuracy indicates whether the utility loss is small, and the attack accuracy indicates how vulnerable the classification model is to MIA.   
We also compare MIA-Pruning with the popular method DP on CIFAR-10 dataset. We followed the same architecture with the four layers CNN classification model in \cite{rahman2018membership} and compare our results with the reported results in ~\cite{rahman2018membership}. For MNIST datasets, we use LeNet-5 as the classification model and implement DP with the noise parameter $\epsilon$ as 6.28. The detailed setting of training can be found in Appendix Section 4.
%When training the classification model, the batch size is set as 64, and the training epoch is 300. The optimizer is Adam, and the learning rate is 0.05. For the MIA-Pruning, we pre-train the classification model for 200 epochs and then we follow the MIA-Pruning process shown in Algorithm \ref{algo}. 

%----------------------------------------------------------
\begin{table}[t]
%\caption{Comparison of classification accuracy and membership attack accuracy on different image datasets between model baseline, MIA-Pruning, and Min-Max Game. }
%\vspace{+0.5cm}
\centering
\scriptsize
\resizebox{\columnwidth}{!}{%
\begin{tabular}{ccc|cc|cccc}
\hline
                                                          & \multicolumn{2}{c|}{Baseline}                                                                                                                                  & \multicolumn{2}{c|}{MIA-Pruning}                                                                                                                & \multicolumn{2}{c}{Min-Max Game}                                                                                                                                                                                    \\ \hline
                                                          & \cellcolor[HTML]{FFFFFF}\begin{tabular}[c]{@{}c@{}}Testing \\ accuracy\end{tabular} & \cellcolor[HTML]{FFCCC9}\begin{tabular}[c]{@{}c@{}}Attack \\ accuracy\end{tabular} & \begin{tabular}[c]{@{}c@{}}Testing \\ accuracy\end{tabular} & \cellcolor[HTML]{FFCCC9}\begin{tabular}[c]{@{}c@{}}Attack \\ accuracy\end{tabular} & \begin{tabular}[c]{@{}c@{}}Testing\\ accuracy\end{tabular} & \cellcolor[HTML]{FFCCC9}\begin{tabular}[c]{@{}c@{}}Attack \\ accuracy\end{tabular}  \\ \hline
\begin{tabular}[c]{@{}c@{}}MNIST-\\ LeNet-5\end{tabular}    & \cellcolor[HTML]{FFFFFF}99.3\%                                                      & \cellcolor[HTML]{FFCCC9}67.00\%                                                    & 99.39\%                                                     & \cellcolor[HTML]{FFCCC9} 53.41\%                                                    & 98.98\%                                                    & \cellcolor[HTML]{FFCCC9}56.00\%                                                   \\ \hline
\begin{tabular}[c]{@{}c@{}}CIFAR-10-\\ VGG16\end{tabular}  & \cellcolor[HTML]{FFFFFF}91.28\%                                                     & \cellcolor[HTML]{FFCCC9}61.99\%                                                    & 91.38\%                                                     & \cellcolor[HTML]{FFCCC9} 59.02\%                                                    & 90.97\%                                                    & \cellcolor[HTML]{FFCCC9}60.35\%                                                  \\ \hline

\begin{tabular}[c]{@{}c@{}}CIFAR-10-\\ MobileNetV2\end{tabular}  & \cellcolor[HTML]{FFFFFF}90.09\%                                                     & \cellcolor[HTML]{FFCCC9}62.75\%                                                    & 86.14\%                                                     & \cellcolor[HTML]{FFCCC9}58.98\%                                                    & 89.71\%                                                    & \cellcolor[HTML]{FFCCC9} 57.91\%                                                   \\ \hline
\begin{tabular}[c]{@{}c@{}}CIFAR-100-\\ VGG16\end{tabular} & \cellcolor[HTML]{FFFFFF}64.84\%                                                     & \cellcolor[HTML]{FFCCC9}67.628\%                                                   & 66.93\%                                                     & \cellcolor[HTML]{FFCCC9}  58.61\%                                                    & 65.71\%                                                    & \cellcolor[HTML]{FFCCC9}68.59\%                                                  \\ \hline

\begin{tabular}[c]{@{}c@{}}CIFAR-100-\\ MobileNetV2\end{tabular}  & \cellcolor[HTML]{FFFFFF}64.14\%                                                     & \cellcolor[HTML]{FFCCC9}66.15\%                                                    & 57.72\%                                                     & \cellcolor[HTML]{FFCCC9} 62.67\%                                                    & 63.36\%                  &
\cellcolor[HTML]{FFCCC9}  62.61\%
\\ \hline
\begin{tabular}[c]{@{}c@{}}CIFAR-100-\\ ResNet-18\end{tabular}  & \cellcolor[HTML]{FFFFFF}73.45\%                                                     & \cellcolor[HTML]{FFCCC9}  69.85\%                                                    & 74.10\%                                                     & \cellcolor[HTML]{FFCCC9}63.50\%                                                    & 69.05\%                                                    & \cellcolor[HTML]{FFCCC9}71.78\%                                                   \\ \hline
\end{tabular}%
}
% \vspace{+0.2 cm}
\caption{Comparison of classification accuracy and membership attack accuracy on different image datasets between model baseline, MIA-Pruning, and Min-Max Game. }
\label{table:acc}

\end{table}
%------------------------------

\subsection{Inference Attack Model}
 
To compare with the Min-Max game, we use the same neural network as the inference attack model as in ~\cite{nasr2018machine} for all experiments except CIFAR-10-CNN in Table~\ref{table:dp}. For CIFAR-10-CNN in Table~\ref{table:dp}, the inference attack model is the same as in ~\cite{rahman2018membership} (see Appendix Section 5). 
%\vspace{+0.1cm}
%-----------------
\begin{table}[h]

%\caption{Comparison of classification accuracy and membership attack accuracy between DP and MIA-Pruning.}
\centering
\scriptsize
\begin{tabular}{ccc|cccc|cc}
\hline
                                                          & \multicolumn{2}{c|}{DP}                                                                                                                                  & \multicolumn{2}{c}{MIA-Pruning}                                                                                                                                                                     \\ \hline
                                                          & \cellcolor[HTML]{FFFFFF}\begin{tabular}[c]{@{}c@{}}Testing\\  accuracy\end{tabular} & \cellcolor[HTML]{FFCCC9}\begin{tabular}[c]{@{}c@{}}Attack\\ accuracy\end{tabular} & \begin{tabular}[c]{@{}c@{}}Testing\\ accuracy\end{tabular} & \cellcolor[HTML]{FFCCC9}\begin{tabular}[c]{@{}c@{}}Attack \\ accuracy\end{tabular}  \\ \hline
\begin{tabular}[c]{@{}c@{}}CIFAR-10-\\ CNN\end{tabular}  & \cellcolor[HTML]{FFFFFF}68.10\%                                                     & \cellcolor[HTML]{FFCCC9}58.30\%                                                    & 75.46\%                                                     & \cellcolor[HTML]{FFCCC9}57.36\%                                                   \\ \hline
\begin{tabular}[c]{@{}c@{}} MNIST-\\ LeNet-5\end{tabular}  & \cellcolor[HTML]{FFFFFF}96.64\%                                                     & \cellcolor[HTML]{FFCCC9}78.54\%                                                    & 99.30\%                                                     & \cellcolor[HTML]{FFCCC9}53.41\%                                                  \\ \hline

\end{tabular}
\caption{Comparison of classification accuracy and membership attack accuracy between DP and MIA-Pruning.}
\label{table:dp}
\end{table}
%--------------------------

%-----------------

\begin{table}[t]
%\vspace{+0.2cm}
%\caption{Comparison of classification accuracy and membership attack accuracy on ImageNet between model baseline and MIA-Pruning.}
\centering
\scriptsize
\begin{tabular}{ccc|cccc|cc}
\hline
                                                          & \multicolumn{2}{c|}{Baseline}                                                                                                                                  & \multicolumn{2}{c}{MIA-Pruning}                                                                                                                                                                     \\ \hline
                                                          & \cellcolor[HTML]{FFFFFF}\begin{tabular}[c]{@{}c@{}}Testing\\  accuracy\end{tabular} & \cellcolor[HTML]{FFCCC9}\begin{tabular}[c]{@{}c@{}}Attack\\ accuracy\end{tabular} & \begin{tabular}[c]{@{}c@{}}Testing\\ accuracy\end{tabular} & \cellcolor[HTML]{FFCCC9}\begin{tabular}[c]{@{}c@{}}Attack \\ accuracy\end{tabular}  \\ \hline

\begin{tabular}[c]{@{}c@{}}ImageNet-\\ MobileNetV2\end{tabular}  & \cellcolor[HTML]{FFFFFF}71.88\%                                                     & \cellcolor[HTML]{FFCCC9}66.90\%                                                    & 68.77\%                                                     & \cellcolor[HTML]{FFCCC9}64.79\%                                                  \\ \hline
\begin{tabular}[c]{@{}c@{}}ImageNet-\\ ResNet-18\end{tabular}  & \cellcolor[HTML]{FFFFFF}69.76\%                                                     & \cellcolor[HTML]{FFCCC9}66.20\%                                                    & 69.30\%                                                     & \cellcolor[HTML]{FFCCC9}61.27\%                                                  \\ \hline
\end{tabular}
% \vspace{+0.2 cm}
\caption{Comparison of classification accuracy and membership attack accuracy on ImageNet between model baseline and MIA-Pruning.}
\label{table:acc2}
\end{table}
%--------------

\subsection{Evaluation Results on MIA-Pruning}

\begin{figure}[h]
	\includegraphics[width=0.49\textwidth]{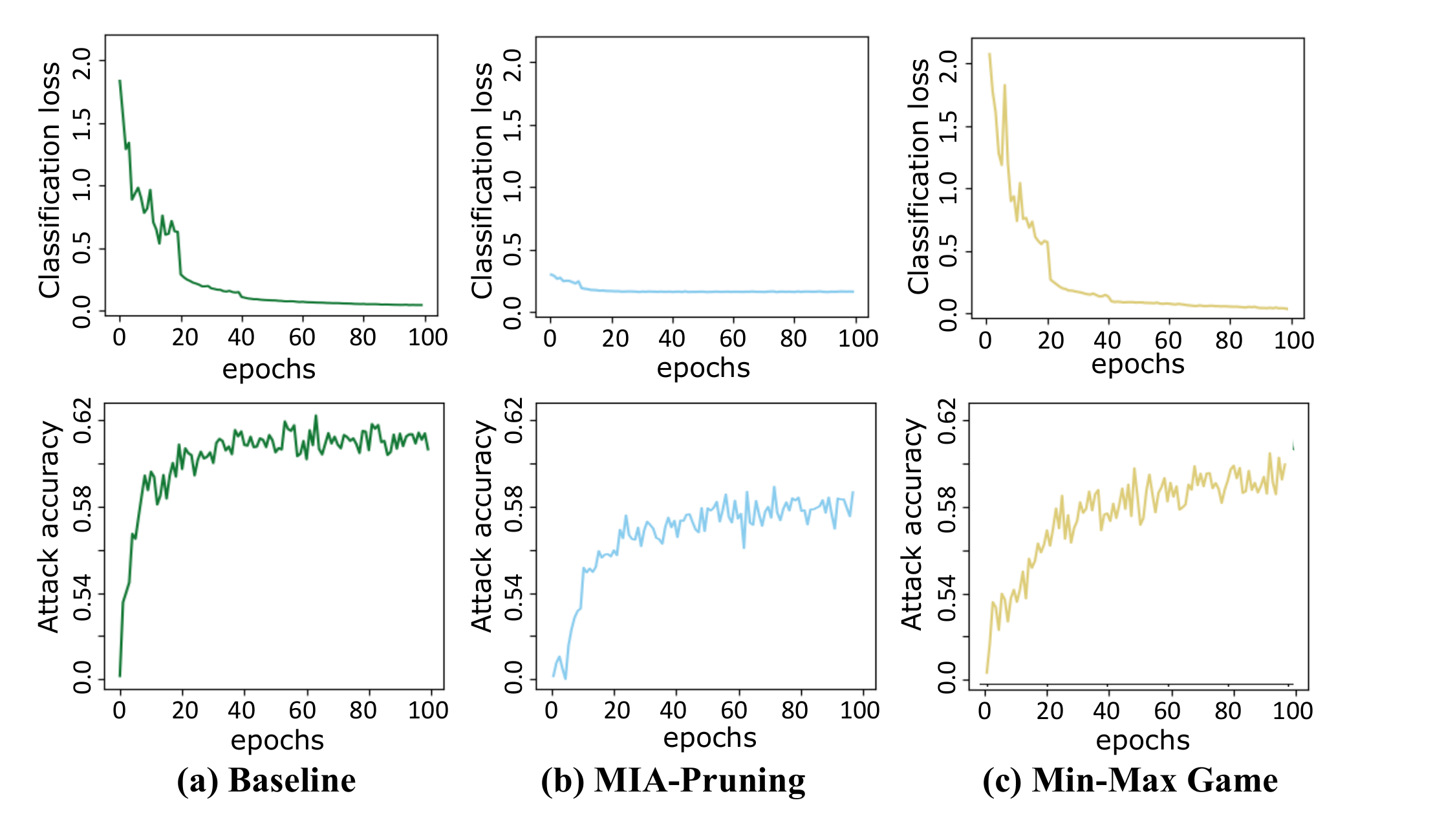}

    \caption{Upper row: the classification loss along with epochs during training classification model. Lower row: membership attack accuracy during training the inference model gain the trained classification model. Classification models baseline, MIA-Pruning, and Min-Max game are shown in (a-c) respectively. }
    \label{fig:loss}
\end{figure}

\subsubsection{MNIST, CIFAR-10 and CIFAR-100}
We compare weight pruning (using MIA-Pruning) and Min-Max game to investigate if weight pruning can constrain the maximum gain $G$ of the inference model, i.e., further reduce attack accuracy. We provide the attack accuracy and testing accuracy of baseline (without defense and pruning), MIA-Pruning, and the Min-Max game as shown in Table~\ref{table:acc}. On MNIST, experimental results demonstrate that for LeNet-5, the attack accuracy using MIA-Pruning is 13.6\% lower than the attack accuracy of baseline and is 2.6\% lower than the attack accuracy of the Min-Max game. From the comparison between DP and MIA-Pruning shown in Table~\ref{table:dp}, MIA-Pruning achieves 25.13\% lower attack accuracy than DP and with 2.66\% higher testing accuracy of the classification model.

On CIFAR-10, the experimental results demonstrate that for VGG16, the attack accuracy using MIA-Pruning is 3\% lower than the baseline attack accuracy and is 1.34\% lower than the attack accuracy of the Min-Max Game. On the other hand, for MobileNetV2, the attack accuracy using MIA-Pruning is 3.77\% lower than the baseline attack accuracy and is close to Min-Max game. As shown in Table~\ref{table:dp}, on a four- layers CNN \cite{rahman2018membership}, MIA-Pruning has 1\% lower attack accuracy than DP, while MIA-Pruning has 7.36\% higher testing accuracy of the classification model than DP.

On CIFAR-100, the experimental results demonstrate that for VGG16, the attack accuracy using MIA-Pruning is 9.1\%  lower than the baseline attack accuracy and is approximately 10\% lower than the Min-Max game. On the other hand, for MobileNetV2, the attack accuracy using MIA-Pruning is 3.48\% lower than the baseline attack accuracy and is close to the Min-Max Game.

%----------------------------------------------------

%--------------------------------------------------------

\begin{figure}[t]
\centering
	\includegraphics[width=0.4\textwidth]{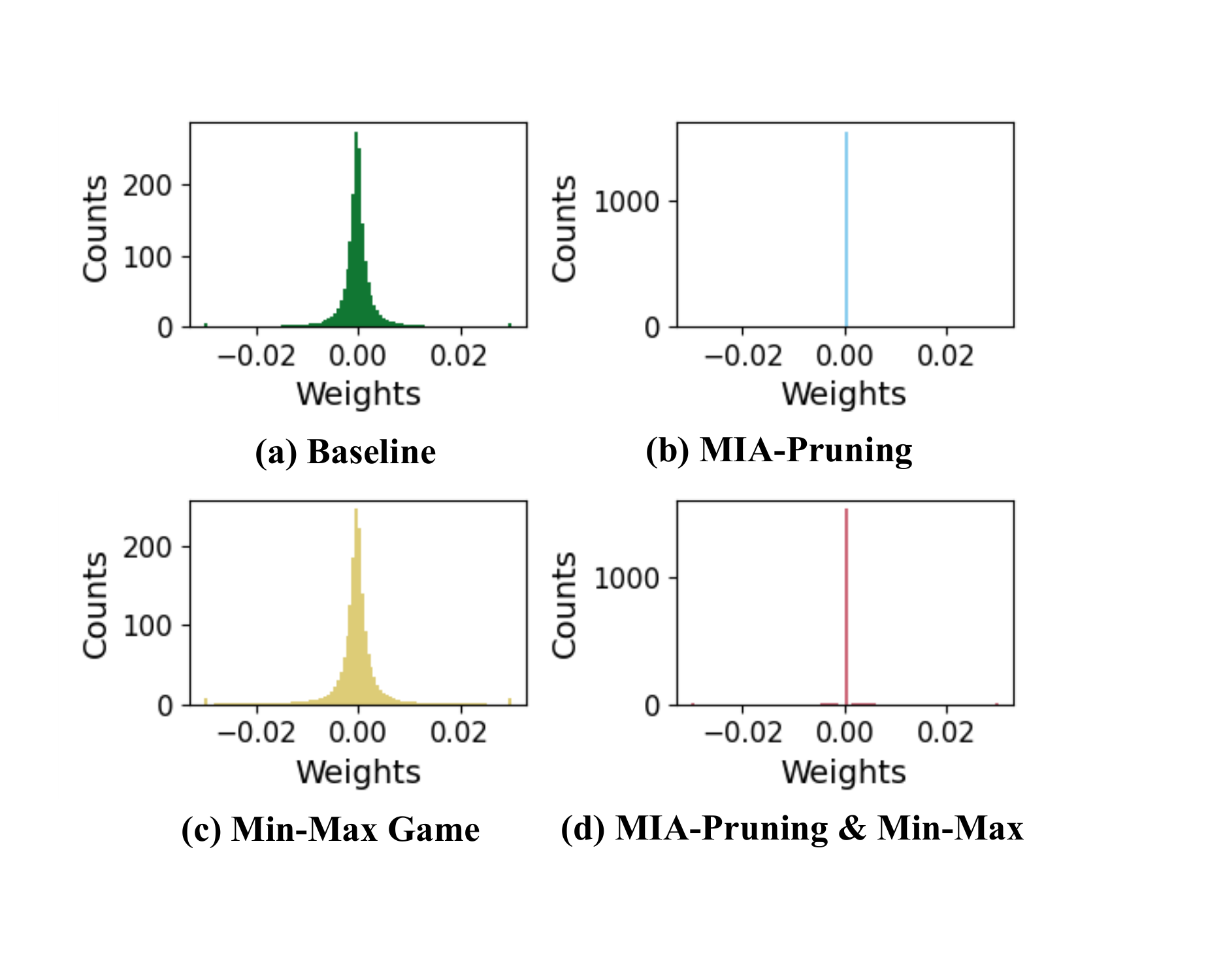}

    \caption{Distribution of weights in VGG16 trained on CIFAR-10 for (a) baseline, (b) MIA-Pruning, (c) Min-Max Game, and (d) MIA-Pruning \& Min-Max. }
    \label{fig:weights}

\end{figure}

The results indicate that using weight pruning can help against MIA, and weight pruning is more effective than using Min-Max Game. On the other hand, weight pruning has significantly less utility cost than DP. And our experiment also shows that DP is hard to achieve privacy-preserving with negligible utility loss. Also, base on the experiment in ~\cite{rahman2018membership}, to achieve the same level of attack accuracy, the test accuracy under the DP method is under 70\% in the best case, 25\% in the worst case on CIFAR-10 by different noise parameter $\epsilon$. In addition, weight pruning brings another benefit shown in Table~\ref{tab:mc_ratio}, i.e., we achieve 15.78X model size reduction for LeNet-5 on MNIST, at least 10.06X model size reduction for on CIFAR-10/CIFAR-100 among VGG16, MobileNetV2, and ResNet-18, which is extremely helpful for deploying DNNs on mobile devices.
Figure~\ref{fig:weights} (a)-(c) show the weight distributions in different classification models from baseline, MIA-Pruning, and Min-Max game. We can observe that the weights after pruning are much less than the baseline model and Min-Max game model (both without pruning).

% \subsubsection{Classification Loss}
Next, we investigate classification loss of baseline (without pruning and defense), MIA-Pruning, and Min-Max Game. Taking CIFAR-10-VGG16 as an example, Figure \ref{fig:loss} shows the classification loss of baseline, MIA-Pruning, and Min-Max Game, respectively, in the upper row. The classification loss of MIA-Pruning converges rapidly in less than 20 epochs. In addition, it has the highest final classification loss when the model is fully trained. In other words, MIA-Pruning prevents overfitting instead of reducing the classification loss on training data arbitrarily low. 
We train the membership inference model based on the predicted outputs of the well-trained classification model. We plot the testing accuracy of membership inference attack during the inference model training process in the lower row in Figure \ref{fig:loss}. The adversary attack accuracy is measured by averaging the adversary's correct determination percentage among all adversary determination for the observed data records \cite{nasr2018machine}.
%\vspace{0.1cm}
%------------------------------
% \usepackage{multirow}
\begin{table}[h]
%\caption{weight pruning ratios for different classification networks. }
%\footnotesize
%\scriptsize
\centering
\resizebox{8cm}{!}{

\begin{tabular}{ccccc} 
\hline
Data                                                & Model      & Weights (\#) & $\begin{matrix} \text{Weights after}\\ \text{prunning (\#)} \end{matrix}$ & $\begin{matrix} \text{Weights}\\ \text{reduction ratio} \end{matrix}$ \\ \hline
\multicolumn{1}{c|}{MNIST}                             & LeNet       & 60 K         & 3.80 K                     & 15.78 $\times$                 \\ \hline
\multicolumn{1}{c|}{\multirow{3}{*}{CIFAR-10/100}} & VGG16       & 13.83 M      & 1.00 M                     & 13.84 $\times$                \\ \cline{2-5} 
\multicolumn{1}{c|}{}                                  & ResNet-18    & 11.17 M      & 1.06 M                     & 10.54 $\times$                \\ \cline{2-5} 
\multicolumn{1}{c|}{}                                  & MobileNetV2 & 3.46 M       & 0.34 M                     & 10.06 $\times$                \\ \hline
\multicolumn{1}{c|}{\multirow{2}{*}{ImageNet}}       
                              & ResNet-18    & 11.17 M     & 3.47 M                     & 3.37 $\times$                    \\ \cline{2-5} 
\multicolumn{1}{c|}{}                                  & MobileNetV2 & 3.46 M       & 1.06 M                     & 3.27 $\times$                  \\ \hline
\end{tabular}
}
%\vspace{+0.1cm}
\caption{weight pruning ratios for different classification networks. }
\label{tab:mc_ratio}
\end{table}
%-----------------

\begin{table}[h]
%\caption{Comparison of classification accuracy and membership attack accuracy between MIA-Pruning and MIA-Pruning \&  Min-Max.  }
\centering
\scriptsize
\begin{tabular}{ccc|cccc|cc}
\hline
                                                          & \multicolumn{2}{c|}{MIA-Pruning}                                                                                                                                  & \multicolumn{2}{c}{MIA--Pruing \& Min-Max}                                                                                                                                                                     \\ \hline
                                                          & \cellcolor[HTML]{FFFFFF}\begin{tabular}[c]{@{}c@{}}Testing\\  accuracy\end{tabular} & \cellcolor[HTML]{FFCCC9}\begin{tabular}[c]{@{}c@{}}Attack\\ accuracy\end{tabular} & \begin{tabular}[c]{@{}c@{}}Testing\\ accuracy\end{tabular} & \cellcolor[HTML]{FFCCC9}\begin{tabular}[c]{@{}c@{}}Attack\\  accuracy\end{tabular}  \\ \hline

\begin{tabular}[c]{@{}c@{}}CIFAR-10-VGG16\end{tabular}  & \cellcolor[HTML]{FFFFFF}91.38\%                                                     & \cellcolor[HTML]{FFCCC9}59.02\%                                                    & 89.19\%                                                     & \cellcolor[HTML]{FFCCC9}54.67\%                                                   \\ \hline
\begin{tabular}[c]{@{}c@{}}CIFAR-100-VGG16\end{tabular}  & \cellcolor[HTML]{FFFFFF}66.93\%                                                     & \cellcolor[HTML]{FFCCC9}58.61\%                                                    & 54.71\%                                                     & \cellcolor[HTML]{FFCCC9}57.65\%                                                  \\ \hline
\begin{tabular}[c]{@{}c@{}}MNIST-LeNet-5\end{tabular}  & \cellcolor[HTML]{FFFFFF}99.39\%                                                     & \cellcolor[HTML]{FFCCC9}53.41\%                                                    & 99.03\%                                                     & \cellcolor[HTML]{FFCCC9}54.00\%                                                  \\ \hline
\end{tabular}
\caption{Comparison of classification accuracy and membership attack accuracy between MIA-Pruning and MIA-Pruning \&  Min-Max.  }

\label{table:acc3}
\end{table}
%--------------

\subsubsection{ImageNet}

The experimental results for ImageNet are shown in Table \ref{table:acc2}, which demonstrates that for MobileNetV2, the attack accuracy using MIA-Pruning is 2.11\% lower than the baseline, then, for ResNet-18, the attack accuracy using pruning is approximately 5\% lower than the baseline. The weight reduction ratio is 3.37 $\times$ for ResNet-18 and 3.27 $\times$ for MobileNetV2 compare with the baseline weights. 

% \vspace{-3 mm}
\subsection{Evaluation Results on MIA-Pruning \& Min-Max}
The experimental result for the MIA-Pruning \& Min-Max is showed in Table \ref{table:acc3}. The experiment results 
% shown in Table \ref{table:acc} 
demonstrate that for CIFAR-10-VGG16, the attack accuracy of Pruning \&Min-Max is 54.67\%, which is 3.03\% lower than the attack accuracy of MIA-Pruning. And for CIFAR-100-VGG16, the attack accuracy of pruning \& Min-Max is 57.65\%, which is 1\% lower than the attack accuracy of MIA-Pruning. For MNIST-LeNet-5, the attack accuracy of pruning \& Min-Max is close to the attack accuracy of MIA-Pruning. Figure~\ref{fig:weights} (d) shows the distribution of weights in classification  models  from MIA-Pruning \& Min-Max. We can also observe that after pruning, the weights are much less than the baseline model.

%--------------------------------------------------------
\begin{table}[]
%\caption{Comparison of classification accuracy on training and testing set and membership attack accuracy between model baseline, MIA-Pruning, Min-Max Game, and MIA-Pruning \& Min-Max.}
\small
\centering
\begin{tabular}{llll}
\hline
                    & \begin{tabular}[c]{@{}l@{}}Training\\ accuracy\end{tabular} & \begin{tabular}[c]{@{}l@{}}Testing\\ accuracy\end{tabular} & \begin{tabular}[c]{@{}l@{}}Attack\\ accuracy\end{tabular} \\ \hline
Baseline  & 99.66\%                                                     & 91.28\%                                                    & 61.99\%                                                   \\ \hline
MIA-Pruning        & 99.22\%                                                     & 91.38\%                                                    & 59.02\%                                                   \\ \hline
Min-Max Game        & 99.18\%                                                     & 90.97\%                                                    & 60.35\%                                                   \\ \hline
MIA-Pruning\&Min-Max & 92.86\%                                                     & 89.19\%                                                    & 54.67\%                                                   \\ \hline
\end{tabular}
\caption{Comparison of classification accuracy on training and testing set and membership attack accuracy between model baseline, MIA-Pruning, Min-Max Game, and MIA-Pruning \& Min-Max.}

\label{tab:train_test_gap}
\end{table}

%----------------------------------------------------
%\begin{figure}[t]
%\centering
%	\includegraphics[width=0.45\textwidth]{model_h.pdf}

%    \caption{Comparison of the predicted probability between baseline, MIA-Pruning, Min-Max Game, and MIA-Pruning \& Min-Max. Upper row: training set; Lower row: testing set. The classifier used is VGG16 and the data is CIFAR-10. Models with defense (i.e., MIA and Min-Max Game) have higher prediction uncertainty.}
%    \label{fig:distribution_model_h}
%\end{figure}

%--------------------------------------------------------

%----------------------------------------------
%\begin{figure}[h]
%\centering
%	\includegraphics[width=0.3\textwidth]{train_test_gap.pdf}
%    \caption{Attack accuracy versus training/non-training accuracy gap of VGG16 on CIFAR-10, which is defined as the difference of classification accuracy between training and non-training data.  }
%    \label{fig:train_test_gap}
%\end{figure}
%--------------------------------------------------------
%\vspace{-0.2cm}
\subsection{MIA-Pruning Analysis}
In general, for the same type of model, the more overfitting the model is, the more vulnerable it is to MIA. The least generic the distribution of training data is, the more information it leaks. MIA-Pruning achieves parameter sparsity by pruning non-critical weights, thus can potentially reduce the overfitting caused by over parameterization. Taking CIFAR-10-VGG16 as an example, we compare the prediction on training data and non-training data between baseline, MIA-Pruning, and Min-Max Game.
%------------------------------------
%Figure \ref{fig:distribution_model_h} shows the predicted probability of class 1 for data belonging to class 1 in the training dataset and testing dataset. 
%-----------------------------------------------
The baseline predicts the highest probability for the correct class in the training data. In other words, it has the highest prediction certainty, or it fits the training data best. However, the baseline has significantly lower prediction certainty on the testing data. The difference of prediction between the training data and testing data can be learned by the inference model to distinguish the membership. Since models with pruning or/and Min-Max Game reduce overfitting, they have lower prediction certainty on the training data, but similar to the prediction certainty on the testing data. The detail of pruning rate settings is in Appendix Section 6. \\ 

To summarize the difference of prediction between training and non-training data quantitatively, we show the MIA accuracy along with the difference of classification accuracy between training and non-training data for each class in CIFAR-10 in
Table \ref{tab:train_test_gap}. The larger the training-non-training accuracy gap is, the higher the membership attack accuracy is.

%Figure \ref{fig:train_test_gap}. We name such difference as training/non-training accuracy gap. As shown in Figure \ref{fig:train_test_gap}, there is a correlation between the train/non-training accuracy gap and the attack accuracy. The larger the training-non-training accuracy gap is, the higher the membership attack accuracy is. Among all the four methods, MIA-Pruning \& Min-Max achieves the smallest train-test accuracy gap and lowest membership inference attack accuracy, therefore providing the highest privacy enhancement. The comparison of overall training accuracy, testing accuracy, and membership inference attack accuracy is illustrated in Table \ref{tab:train_test_gap}, which conveys similar messages as Figure \ref{fig:train_test_gap}. \\

%\vspace{-0.5cm}
%----------------------------------------------------
\section{Conclusion}
In this work, we jointly formulate weight pruning and MIA as MIA-Pruning and provide an algorithm to solve the problem.
We theoretically analyze and conclude that our proposed algorithm can protect the information privacy from MIA and achieves competitive accuracy with the original DNNs. And we evaluate our method on LeNet-5, VGG16, MobileNetV2, ResNet-18 on different datasets including MNIST, CIFAR-10, CIFAR-100, and ImageNet. From experimental results, we see weight pruning can significantly reduce the information leakage from MIA.
% Compared with  Min-Max game, we show that our MCMIA model can reduce the information leakage from MIA. 

Our proposed method outperforms DP on MIA.
Compared with our MIA-Pruning, our MIA-Pruning \& Min-Max game can achieve the lowest attack accuracy so that it maximally enhance DNN model privacy.
Thanks to the hardware-friendly characteristic of weight pruning (reducing weight storage and computational operation), our proposed MIA-Pruning is
very helpful for deploying DNNs on mobile devices. 
We hope our proposed method will shed some light on the increasing membership privacy concerns when applying DNNs on user-sensitive data such as business and medical datasets on mobile devices.

\vspace{-0.2cm}
\section*{Acknowledgements}
This work has been supported in part by the following NSF grants: 1743418 and 1843025. This work originated from the course project of CSE5095 Advances in Deep Learning taught by Prof. Caiwen Ding. We also thank Dan Guo for many helpful discussions.

\begin{small}
%%
%% The next two lines define the bibliography style to be used, and
%% the bibliography file.
\bibliographystyle{named}
\bibliography{ijcai21}

\begin{thebibliography}{}

\bibitem[\protect\citeauthoryear{Abadi \bgroup \em et al.\egroup
  }{2016}]{abadi2016deep}
Martin Abadi, Andy Chu, Ian Goodfellow, H~Brendan McMahan, Ilya Mironov, Kunal
  Talwar, and Li~Zhang.
\newblock Deep learning with differential privacy.
\newblock In {\em Proceedings of the 2016 ACM SIGSAC conference on computer and
  communications security}, 2016.

\bibitem[\protect\citeauthoryear{Alvim \bgroup \em et al.\egroup
  }{2017}]{alvim2017information}
M{\'a}rio~S Alvim, Konstantinos Chatzikokolakis, Yusuke Kawamoto, and Catuscia
  Palamidessi.
\newblock Information leakage games.
\newblock In {\em International Conference on Decision and Game Theory for
  Security}, pages 437--457. Springer, 2017.

\bibitem[\protect\citeauthoryear{Blalock \bgroup \em et al.\egroup
  }{2020}]{blalock2020state}
Davis Blalock, Jose Javier~Gonzalez Ortiz, Jonathan Frankle, and John Guttag.
\newblock What is the state of neural network pruning?
\newblock {\em arXiv preprint arXiv:2003.03033}, 2020.

\bibitem[\protect\citeauthoryear{Boyd \bgroup \em et al.\egroup
  }{2011}]{boyd2011distributed}
Stephen Boyd, Neal Parikh, Eric Chu, Borja Peleato, and Jonathan Eckstein.
\newblock Distributed optimization and statistical learning via the alternating
  direction method of multipliers.
\newblock {\em Foundations and Trends{\textregistered} in Machine learning},
  3(1):1--122, 2011.

\bibitem[\protect\citeauthoryear{Chaudhuri \bgroup \em et al.\egroup
  }{2011}]{chaudhuri2011differentially}
Kamalika Chaudhuri, Claire Monteleoni, and Anand~D Sarwate.
\newblock Differentially private empirical risk minimization.
\newblock {\em Journal of Machine Learning Research}, 12(Mar):1069--1109, 2011.

\bibitem[\protect\citeauthoryear{Frankle and Carbin}{2018}]{frankle2018lottery}
Jonathan Frankle and Michael Carbin.
\newblock The lottery ticket hypothesis: Finding sparse, trainable neural
  networks.
\newblock {\em arXiv preprint arXiv:1803.03635}, 2018.

\bibitem[\protect\citeauthoryear{Gollob}{2015}]{gollob2015microsoft}
David Gollob.
\newblock {\em Microsoft Azure-Planning, Deploying, and Managing Your Data
  Center in the}.
\newblock Springer-verlag Berlin And Hei, 2015.

\bibitem[\protect\citeauthoryear{Goodfellow \bgroup \em et al.\egroup
  }{2014}]{goodfellow2014generative}
Ian Goodfellow, Jean Pouget-Abadie, Mehdi Mirza, Bing Xu, David Warde-Farley,
  Sherjil Ozair, Aaron Courville, and Yoshua Bengio.
\newblock Generative adversarial nets.
\newblock In {\em Advances in neural information processing systems}, pages
  2672--2680, 2014.

\bibitem[\protect\citeauthoryear{He \bgroup \em et al.\egroup
  }{2016}]{he2016deep}
Kaiming He, Xiangyu Zhang, Shaoqing Ren, and Jian Sun.
\newblock Deep residual learning for image recognition.
\newblock In {\em Proceedings of the IEEE Conference on Computer Vision and
  Pattern Recognition}, pages 770--778, 2016.

\bibitem[\protect\citeauthoryear{Hinton \bgroup \em et al.\egroup
  }{2012}]{hinton2012deep}
Geoffrey Hinton, Li~Deng, Dong Yu, George~E Dahl, Abdel-rahman Mohamed, Navdeep
  Jaitly, Andrew Senior, Vincent Vanhoucke, Patrick Nguyen, Tara~N Sainath, and
  Brian Kingsbury.
\newblock Deep neural networks for acoustic modeling in speech recognition: The
  shared views of four research groups.
\newblock {\em IEEE Signal Processing Magazine}, 2012.

\bibitem[\protect\citeauthoryear{Hong \bgroup \em et al.\egroup
  }{2016}]{hong2016convergence}
Mingyi Hong, Zhi-Quan Luo, and Meisam Razaviyayn.
\newblock Convergence analysis of alternating direction method of multipliers
  for a family of nonconvex problems.
\newblock {\em SIAM Journal on Optimization}, 26(1):337--364, 2016.

\bibitem[\protect\citeauthoryear{Iyengar \bgroup \em et al.\egroup
  }{2019}]{iyengar2019towards}
Roger Iyengar, Joseph~P Near, Dawn Song, Om~Thakkar, Abhradeep Thakurta, and
  Lun Wang.
\newblock Towards practical differentially private convex optimization.
\newblock In {\em 2019 IEEE Symposium on Security and Privacy (SP)}. IEEE,
  2019.

\bibitem[\protect\citeauthoryear{Jia \bgroup \em et al.\egroup
  }{2019}]{jia2019memguard}
Jinyuan Jia, Ahmed Salem, Michael Backes, Yang Zhang, and Neil~Zhenqiang Gong.
\newblock Memguard: Defending against black-box membership inference attacks
  via adversarial examples.
\newblock In {\em Proceedings of the 2019 ACM SIGSAC Conference on Computer and
  Communications Security}, 2019.

\bibitem[\protect\citeauthoryear{Krizhevsky \bgroup \em et al.\egroup
  }{2012}]{krizhevsky2012imagenet}
Alex Krizhevsky, Ilya Sutskever, and Geoffrey~E Hinton.
\newblock Imagenet classification with deep convolutional neural networks.
\newblock In {\em Advances in neural information processing systems}, 2012.

\bibitem[\protect\citeauthoryear{Kurniawan}{2018}]{kurniawan2018learning}
Agus Kurniawan.
\newblock {\em Learning AWS IoT: Effectively manage connected devices on the
  AWS cloud using services such as AWS Greengrass, AWS button, predictive
  analytics and machine learning}.
\newblock Packt Publishing Ltd, 2018.

\bibitem[\protect\citeauthoryear{Liu \bgroup \em et al.\egroup
  }{2018}]{liu2018zeroth}
Sijia Liu, Jie Chen, Pin-Yu Chen, and Alfred Hero.
\newblock Zeroth-order online alternating direction method of multipliers:
  Convergence analysis and applications.
\newblock In {\em International Conference on Artificial Intelligence and
  Statistics}, pages 288--297, 2018.

\bibitem[\protect\citeauthoryear{Lueker}{1998}]{lueker1998exponentially}
George~S Lueker.
\newblock Exponentially small bounds on the expected optimum of the partition
  and subset sum problems.
\newblock {\em Random Structures \& Algorithms}, 12(1):51--62, 1998.

\bibitem[\protect\citeauthoryear{Nasr \bgroup \em et al.\egroup
  }{2018}]{nasr2018machine}
Milad Nasr, Reza Shokri, and Amir Houmansadr.
\newblock Machine learning with membership privacy using adversarial
  regularization.
\newblock In {\em Proceedings of the 2018 ACM SIGSAC Conference on Computer and
  Communications Security}, pages 634--646, 2018.

\bibitem[\protect\citeauthoryear{Ouyang \bgroup \em et al.\egroup
  }{2013}]{ouyang2013stochastic}
Hua Ouyang, Niao He, Long Tran, and Alexander Gray.
\newblock Stochastic alternating direction method of multipliers.
\newblock In {\em International Conference on Machine Learning}, pages 80--88,
  2013.

\bibitem[\protect\citeauthoryear{Pan \bgroup \em et al.\egroup
  }{2020}]{9127461}
Haihong Pan, Zaijun Pang, Yaowei Wang, Yijue Wang, and Lin Chen.
\newblock A new image recognition and classification method combining transfer
  learning algorithm and mobilenet model for welding defects.
\newblock {\em IEEE Access}, 2020.

\bibitem[\protect\citeauthoryear{Rahman \bgroup \em et al.\egroup
  }{2018}]{rahman2018membership}
Md~Atiqur Rahman, Tanzila Rahman, Robert Lagani{\`e}re, Noman Mohammed, and
  Yang Wang.
\newblock Membership inference attack against differentially private deep
  learning model.
\newblock {\em Transactions on Data Privacy}, 11(1):61--79, 2018.

\bibitem[\protect\citeauthoryear{Ramanujan \bgroup \em et al.\egroup
  }{2020}]{ramanujan2020s}
Vivek Ramanujan, Mitchell Wortsman, Aniruddha Kembhavi, Ali Farhadi, and
  Mohammad Rastegari.
\newblock What's hidden in a randomly weighted neural network?
\newblock In {\em Proceedings of the IEEE/CVF Conference on Computer Vision and
  Pattern Recognition}, pages 11893--11902, 2020.

\bibitem[\protect\citeauthoryear{Ravulavaru}{2018}]{ravulavaru2018google}
Arvind Ravulavaru.
\newblock {\em Google Cloud AI Services Quick Start Guide: Build Intelligent
  Applications with Google Cloud AI Services}.
\newblock Packt Publishing Ltd, 2018.

\bibitem[\protect\citeauthoryear{Ribeiro \bgroup \em et al.\egroup
  }{2015}]{ribeiro2015mlaas}
Mauro Ribeiro, Katarina Grolinger, and Miriam~AM Capretz.
\newblock Mlaas: Machine learning as a service.
\newblock In {\em 2015 IEEE 14th International Conference on Machine Learning
  and Applications (ICMLA)}, pages 896--902. IEEE, 2015.

\bibitem[\protect\citeauthoryear{Salem \bgroup \em et al.\egroup
  }{2018}]{salem2018ml}
Ahmed Salem, Yang Zhang, Mathias Humbert, Pascal Berrang, Mario Fritz, and
  Michael Backes.
\newblock Ml-leaks: Model and data independent membership inference attacks and
  defenses on machine learning models.
\newblock {\em arXiv preprint arXiv:1806.01246}, 2018.

\bibitem[\protect\citeauthoryear{Shokri \bgroup \em et al.\egroup
  }{2012}]{shokri2012protecting}
Reza Shokri, George Theodorakopoulos, Carmela Troncoso, Jean-Pierre Hubaux, and
  Jean-Yves Le~Boudec.
\newblock Protecting location privacy: optimal strategy against localization
  attacks.
\newblock In {\em Proceedings of the 2012 ACM conference on Computer and
  communications security}, 2012.

\bibitem[\protect\citeauthoryear{Shokri \bgroup \em et al.\egroup
  }{2017}]{shokri2017membership}
Reza Shokri, Marco Stronati, Congzheng Song, and Vitaly Shmatikov.
\newblock Membership inference attacks against machine learning models.
\newblock In {\em 2017 IEEE Symposium on Security and Privacy (SP)}, pages
  3--18. IEEE, 2017.

\bibitem[\protect\citeauthoryear{Shokri}{2015}]{shokri2015privacy}
Reza Shokri.
\newblock Privacy games: Optimal user-centric data obfuscation.
\newblock {\em Proceedings on Privacy Enhancing Technologies}, 2015.

\bibitem[\protect\citeauthoryear{Truex \bgroup \em et al.\egroup
  }{2019}]{truex2019demystifying}
Stacey Truex, Ling Liu, Mehmet~Emre Gursoy, Lei Yu, and Wenqi Wei.
\newblock Demystifying membership inference attacks in machine learning as a
  service.
\newblock {\em IEEE Transactions on Services Computing}, 2019.

\bibitem[\protect\citeauthoryear{Vaswani \bgroup \em et al.\egroup
  }{2017}]{vaswani2017attention}
Ashish Vaswani, Noam Shazeer, Niki Parmar, Jakob Uszkoreit, Llion Jones,
  Aidan~N Gomez, {\L}ukasz Kaiser, and Illia Polosukhin.
\newblock Attention is all you need.
\newblock In {\em Advances in neural information processing systems}, 2017.

\bibitem[\protect\citeauthoryear{Zhang \bgroup \em et al.\egroup
  }{2018}]{zhang2018systematic}
Tianyun Zhang, Shaokai Ye, Kaiqi Zhang, Jian Tang, Wujie Wen, Makan Fardad, and
  Yanzhi Wang.
\newblock A systematic dnn weight pruning framework using alternating direction
  method of multipliers.
\newblock In {\em Proceedings of the European Conference on Computer Vision
  (ECCV)}, pages 184--199, 2018.

\bibitem[\protect\citeauthoryear{Zhang \bgroup \em et al.\egroup
  }{2019}]{zhang2019predictive}
Xueru Zhang, Chunan Huang, Mingyan Liu, Anna Stefanopoulou, and Tulga Ersal.
\newblock Predictive cruise control with private vehicle-to-vehicle
  communication for improving fuel consumption and emissions.
\newblock {\em IEEE Communications Magazine}, 2019.

\bibitem[\protect\citeauthoryear{Zhou \bgroup \em et al.\egroup
  }{2019}]{zhou2019deconstructing}
Hattie Zhou, Janice Lan, Rosanne Liu, and Jason Yosinski.
\newblock Deconstructing lottery tickets: Zeros, signs, and the supermask.
\newblock In {\em Advances in Neural Information Processing Systems}, pages
  3597--3607, 2019.

\end{thebibliography}
\end{small}
%%
%% If your work has an appendix, this is the place to put it.
\appendix

\section{Appendix}
In the appendix, we first introduce the unified problem reformulation of MIA-Pruning, then describe the solution strategy for weight pruning, namely ADMM. We also included the details of experimental settings: the training of classification model, the inference attack model, and the pruning rate.

\subsection{SECTION 1: The Proof of Theorem 1}

\subsubsection{Preliminaries and notations} 
%\cd{They are so many variabls without definition. Let me reemphasize here, you are NOT writing a course project. This is NOT the way of doing Scietific writing. Take a look at any math textbook to see how other people write papers.}
In this part, we introduce some notations we will use in the following analysis. We define the target network $f(x)$ as:
\begin{equation}
    f(x)=W_{n}^f\sigma (W_{n-1}^f...(\sigma(W_1^f x))
    \label{def_f}
\end{equation}
and we define the original network $g(x)$ as:
\begin{equation}
    g(x)= W_{2n}^g\sigma(W_{2n-1}^g...\sigma(W_1^g(x))
    \label{def_g}
\end{equation}
where $W_i^f$,$W_j^g$ is the randomized weight matrix at $i$-th layer of $f$ and $j$-th layer of $g(x)$. And $\sigma(\cdot)$ is the activation function.\\
A pruned network $\hat{g}(x)$ can be presented as :
\begin{equation}
    \hat{g}(x) = (P_{2n} \odot W_{2n}^g)\sigma(P_{2n-1} \odot W_{2n-1}^g)...\sigma(P_{1} \odot W_{1}^gx)
\end{equation}
Where $P_l$ is the prunning matrix in $l$-th layer.

we analysis the objective from simple to complex. We will start from a pruning simple linear function with one variable. Then we consider about pruning a simple ReLU network. Furthermore, we analysis the pruning from a neuron and a layer. Finally,  we give the analysis of our main objective. 

\subsubsection{Analysis on Simple linear Network}
In this case, $f(x) = w \cdot x$ , and $g(x)=  \left (  \sum_{i=1}^{d}w_i\right )x$.

{\bf Theorem 2.} {\it Let $W^*_1,...,W^*_n$ belongs to i.i.d. Uniform distribution over [-1,1], where $n \geq Clog\frac{2}{\delta}$ ,where $\delta \leq min\{1,\epsilon\}$. Then, with probability at least 1-$\delta$, we have \\
\begin{equation}
\begin{aligned}
    &\exists S \subset \{1,2,...,n\}, \forall W \in [-0.5,0.5],\\
    &s.t \left | W-\sum_{i \in S}W^*_i \right | \leq \epsilon 
\end{aligned}
\end{equation}
}

Lueker et al.\cite{lueker1998exponentially} proposed this theorem and have gave a proof.

\subsubsection{Analysis on Simple ReLU Networks}
In this case, $f(x) = w \cdot x$, $g(x) = \mathbf{u}\sigma (\mathbf{w}^g x)$ .
because $\sigma$ is ReLU activation function, we have $w =\sigma (w) - \sigma(-w)$. So that the a single ReLU neuron can be written as:

\begin{equation}
\begin{aligned}
x^*  \mapsto \sigma \left ( wx \right ) = \sigma\left ( \sigma(wx)  - \sigma(-wx) \right)
\label{neuron_o} 
\end{aligned}
\end{equation}

On the other hand, this neuron can be present by a width m two layer network with a pruning matrix $p^*$ for the first layer as:
\begin{equation}
    x^*  \mapsto  \mathbf{u} \sigma \left ( \mathbf{p}\odot  \mathbf{w}^g x\right ) 
    \label{neuron_f}
\end{equation}

we define $\mathbf{w^+}=max\{\mathbf{0},\mathbf{w}\}$,$\mathbf{w^-}=min\{\mathbf{0},\mathbf{w}\}$, $\mathbf{w^+}+\mathbf{w^-}=\mathbf{w}^g$. 
Combine Eq. \ref{neuron_o} and \ref{neuron_f} we have:
\begin{equation}
    x^*  \mapsto  \mathbf{u} \sigma\left( \sigma \left ( \mathbf{p}\odot  \mathbf{w^+}x\right )-\sigma \left ( \mathbf{p}\odot  \mathbf{-w^-}x\right )  \right ) 
\end{equation}
Base on Theorem 2, when $n \geq Clog4/\epsilon$, there exist a pattern of $\mathbf{w}$, such that, with probability $1-\epsilon/2$, 
\begin{equation}
    \forall w^f \in [0,1],  \exists  p \in {0,1}^n, s.t.  \left | w^f- \mathbf{u} \sigma( \mathbf{p} \odot \mathbf{w^+}) \right | < \epsilon/2
    \label{wplus}
\end{equation}

Similarly, we have 
$\mathbf{w}$, such that, with probability $1-\epsilon/2$, 
\begin{equation}
    \forall w^f \in [0,1],  \exists  p \in {0,1}^n, s.t.  \left | w^f- \mathbf{u} \sigma( \mathbf{p} \odot \mathbf{w^-}) \right | < \epsilon/2
    \label{wminus}
\end{equation}

so Consider Eq.\ref{wplus} and \ref{wminus}, we have:

\begin{equation}
\begin{split}
&\text{sup} \left | w^fx- \mathbf{u} \sigma( \mathbf{p} \odot \mathbf{w}x) \right | \\
& \leq \left | \sigma(w^f)x-\sigma(-w^f)x- \mathbf{u} \sigma( \mathbf{p} \odot \mathbf{w^+}x)- \mathbf{u} \sigma( \mathbf{p} \odot \mathbf{w^-}x) \right | \\
& \leq  \text{sup} \left | \sigma (w^f)x- \mathbf{u} \sigma( \mathbf{p} \odot \mathbf{w^+}x) \right |+ \\
& \text{sup} \left | \sigma (w^f)x- \mathbf{u} \sigma( \mathbf{p} \odot \mathbf{w^-}x) \right | \\
&\leq \epsilon/2 + \epsilon/2\\ 
&\leq \epsilon
\end{split}
\label{conv_relu}
\end{equation}

\subsection {Analysis on a Neuron }
In this case, $f(x)= \mathbf{w}^f\mathbf{x}$, $g(x)=\mathbf{u} \sigma(\mathbf{wx})$ and $\hat{g}(x)=\mathbf{u} \sigma( \mathbf{p} \odot \mathbf{wx})$
\begin{equation}
\begin{split}
\text{sup}
\left | \mathbf{w}^f\mathbf{x}- \mathbf{u} \sigma( \mathbf{p} \odot \mathbf{wx}) \right | \\
\leq \text{sup} \left | \sum_{i=1}^{m}\left (  w_i^fx_i- \mathbf{u}_i \sigma( \mathbf{p}_i \odot \mathbf{w}_ix_i)\right ) \right | \\ 
\leq \text{sup}\sum_{i=1}^{m} \left | w_i^fx_i- \mathbf{u}_i \sigma( \mathbf{p}_i \odot \mathbf{w}_ix_i) \right | \\ 
\leq \sum_{i=1}^{m} \text{sup} \left | w_i^fx_i- \mathbf{u}_i \sigma( \mathbf{p}_i \odot \mathbf{w}_ix_i) \right | \\ 
\leq  m \cdot  \frac{\epsilon}{m}\\
\leq \epsilon
\end{split}
\label{conv_neuron}
\end{equation}

\subsection{Analysis on a Layer}
In this case,$f(x)= \mathbf{W}^f\mathbf{x}$, and $g(x)=\mathbf{u} \sigma( \mathbf{W}^g\mathbf{x})$, and $\hat{g}(x)=\mathbf{u} \sigma( \mathbf{p} \odot \mathbf{W}^g\mathbf{x})$
\begin{equation}
\begin{split}
&\text{sup}
\left | \mathbf{W}^f\mathbf{x}- \mathbf{u} \sigma( \mathbf{p} \odot \mathbf{W}^g\mathbf{x}) \right | \\
&\leq \text{sup} \left | \sum_{j=1}^{k}\sum_{i=1}^{m}\left (  w_{j,i}^fx_i- \mathbf{u}_i \sigma( \mathbf{p}_{j,i} \odot \mathbf{w}_{j,i}x_i)\right ) \right | \\ 
&\leq \text{sup}\sum_{j=1}^{k}\sum_{i=1}^{m} \left | w_{j,i}^fx_i- \mathbf{u}_i \sigma( \mathbf{p}_{j,i} \odot \mathbf{w}_{j,i}x_i) \right | \\ 
&\leq \sum_{j=1}^{k}\sum_{i=1}^{m} \text{sup} \left | w_{j,i}^fx_i- \mathbf{u}_i \sigma( \mathbf{p}_{j,i} \odot \mathbf{w}_{j,i}x_i) \right | \\ 
&\leq  k \cdot m  \cdot \frac{\epsilon}{mk}\\
&\leq \epsilon
\end{split}
\label{conv_layer}
\end{equation}

\subsection {The analysis in general}
For general case , $f(x)$ is defined as Eq.\ref{def_f},  $g(x)$ is defined as Eq.\ref{def_g}. so with the probability over $1-\epsilon$,
we have:
\begin{equation}
\begin{split}
& \text{sup} \left \| f(x)-\hat g(x) \right \| \\ 
& =\left \| \mathbf{W}_n \mathbf{x}_n - \mathbf{P}_{2n} \odot \mathbf{W}^g_{2n}\mathbf{x}^g_n \sigma(\mathbf{P}_{2n-1} \odot \mathbf{x}^g_{2n-1}) \right \|  \\
&\leq  \left \| \mathbf{W}_n\mathbf{x}_n -  \mathbf{W}_n\mathbf{x}_n^g \right \| + \\ 
& \left \| \mathbf{W}_n\mathbf{x}_n^g -  \mathbf{P}_{2n} \odot \mathbf{W}^g_{2n}\mathbf{x}^g_n \sigma(\mathbf{P}_{2n-1} \odot \mathbf{x}^g_{2n-1}) \right \|  \\
&\leq \left \| \mathbf{x}_n-\mathbf{x}_n ^g \right \| + \\
&\left \| \mathbf{W}_n\mathbf{x}_n^g -  \mathbf{P}_{2n} \odot \mathbf{W}^g_{2n}\mathbf{x}^g_n \sigma(\mathbf{P}_{2n-1} \odot \mathbf{x}^g_{2n-1}) \right \|  \\
&\leq \epsilon/2+ \epsilon/2\\
&\leq \epsilon
\end{split}
\end{equation}

\subsection{SECTION 2: Unified Problem Reformulation of MIA-pruning}
Comparing Equation 2 and 3 in the main text, the optimization problem of MIA-Pruning can be considered as a special case of MIA-Pruning+Min-Max where $\gamma = 0$. Since ${n_i}$ ranges from 0 to 1 and has a relatively smaller search space, we obtain the optimal ${n_i}$ by simply iterating different sets of ${n_i}$ and choosing the smallest $G_f(f_A^*)$. Therefore, we can rewrite the problem in a unified form and in an equivalent form without constrains:
\begin{equation}
\begin{aligned}
    &  \underset{\{{\bf{W}}_{i}\},\{{\bf{b}}_{i}\}}{\text{min}} \quad
    \mathcal{L} (f(\{{\bf{W}}_i\},\{{\bf{b}}_i\};x),y)\\
    &\text{  }+ \gamma \text{ } \underset{f_A} {\text{max }} G_f (f_A(f(\{{\bf{W}}_{i}\},\{{\bf{b}}_{i}\};x),y))+ \sum_{i=1}^{N} g_i(\mathbf{W}_i) 
\end{aligned}
\label{eq:unif_p}
\end{equation}
where $g_i(\mathbf{W}_i)$ is a indicator function of the constrain $card(\mathbf{W}_i) \leq n_i$. If $card(\mathbf{W}_i) \leq n_i$ holds, $g_i(\mathbf{W}_i) = 0$, otherwise $g_i(\mathbf{W}_i) = + \infty$.

\section{SECTION 3: Solution Strategy: ADMM}
\medskip \noindent{\it ADMM-based DNN weight pruning:} 
Considering an optimization problem $\min_{{\bf{x}}} f(\bf{x})$ with combinatorial constraints, which is difficult to solve directly using optimization tools \cite{zhang2018systematic}.
By using ADMM~\cite{boyd2011distributed}, the $\min_{{\bf{x}}} f(\bf{x})$ problem can be decomposed into two subproblems on $\bf{x}$ and $\bf{z}$ (auxiliary variable), i.e., the first subproblem derives $\bf{x}$ given $\bf{z}$: $\min_{\bf{x}} f({\bf{x}})+q_1(\bf{x}|\bf{z})$; the second subproblem derives $\bf{z}$ given $\bf{x}$: $\min_{\bf{z}} I({\bf{z}})+q_2(\bf{z}|\bf{x})$. 
Both $q_1$ and $q_2$ are quadratic functions. In such way, the two subproblems could be solved separately and iteratively until convergence. 
Originally, ADMM  is used to accelerate the convergence of convex optimization problems and enable distributed optimization, where the optimality and fast convergence rate have been proven~\cite{ouyang2013stochastic,boyd2011distributed}. One special property of ADMM is that it can effectively deal with a subset of combinatorial constraints and yields optimal (or at least high quality) solutions \cite{hong2016convergence,liu2018zeroth}. The related constraints in DNN weight pruning belong to this subset of combinatorial constraints. Therefore ADMM is applicable to DNN mode compression.

In order to solve the constrained optimization problem, we use ADMM method to decompose it into simpler sub-problems. First we rewrite the format of ADMM \cite{zhang2018systematic}:
\begin{equation}
\begin{aligned}
    \underset{\{{\bf{W}}_{i}\},\{{\bf{b}}_{i}\}}{\text{argmin}} \quad
    &\mathcal{L} (f(\{{\bf{W}}_i\},\{{\bf{b}}_i\};x),y)\\
    &+ \gamma \text{ } \underset{f_A} {\text{max }} G_f (f_A(f(\{{\bf{W}}_{i}\},\{{\bf{b}}_{i}\};x),y))\\
    &+ \sum_{i=1}^{N} g_i(\mathbf{Z}_i)\\
    &\text{s.t.} \quad \mathbf{W}_i = \mathbf{Z}_i \text{, i = 1, ..., N}
\end{aligned}
\label{eq:admm}
\end{equation}
For simplicity, we denote the first two parts of the minimization objective in the above equation as $\mathcal{L}_s(\{{\bf{W}}_i\},\{{\bf{b}}_i\})$. The augmented Lagrangian solution for the above problem can be written as: 
\begin{equation}
\begin{aligned}
 &L_{\lambda} \big( \{{\bf{W}}_{i} \}, \{{\bf{b}}_{i} \}, \{{\bf{Z}}_{i} \}, \{\mathbf{U}_i \} \big)\\
 &~~~=\mathcal{L}_s(\{{\bf{W}}_i\},\{{\bf{b}}_i\}) +\sum_{i=1}^{N} g_i(\mathbf{Z}_i)\\
 & \quad + \sum_{i=1}^{N} \frac{\lambda_i}{2} \parallel {\bf{W}}_{i}-{\bf{Z}}_{i} +\mathbf{U}_i \parallel_F^2
 + \sum_{i=1}^{N} \frac{\lambda_i}{2} \parallel \mathbf{U}_i \parallel_F^2
 \end{aligned}
\end{equation}
where ${\bf{U}}_{i}=(1/\lambda_i){\bf{\Lambda}}_{i}$. ${\bf{\Lambda}}_{i}$ is the Lagrange dual variable, $\lambda_i$ is penalty parameters, and $\parallel \cdot \parallel_F^2$ is the Frobenius norm.\\ 
The ADMM process updates the parameters repeatedly as follows: 
\begin{subequations}
\begin{align}
%renew h
%h^{k+1}&:=\mathop{\arg\max}_{{h}^{k}}  G(h^k({\bf{W}}_i,{\bf{b}}_i))
%\\
%renew w , b
&\{{\bf{W}}_{i}^{k+1},{\bf{b}}_{i}^{k+1}\}=\mathop{\arg\min}_{ \{{\bf{W}}_{i}\}, \{{\bf{b}}_{i} \}}\ L_{\lambda} {\big( \{{\bf{W}}_{i}\}, \{{\bf{b}}_{i} \}, \{{\bf{Z}}_{i}^{k}\}, \{{\bf{U}}_{i}^{k}\} \big)} \label{eq:sub1}\\
%renew z
&\{{\bf{Z}}_{i}^{k+1} \}=\mathop{\arg\min}_{ \{{\bf{Z}}_{i} \}}\ L_{\lambda} \big( \{{\bf{W}}_{i}^{k+1} \}, \{{\bf{b}}_{i}^{k+1} \},\{{\bf{Z}}_{i} \}, \{{\bf{U}}_{i}^{k} \} \big)  
\label{eq:sub2}\\
%renew u
&{\bf{U}}_{i}^{k+1}={\bf{U}}_{i}^{k}+{\bf{W}}_{i}^{k+1}-{\bf{Z}}_{i}^{k+1}
\label{eq:sub3}
\end{align}
\end{subequations}
The sub-problem of updating $\{\mathbf{W_i}, \mathbf{b}_i\}$ in Equation \ref{eq:sub1} is easy to solve:
\begin{equation}
\begin{aligned}
\mathbf{W}_i^{k+1} &= \frac{\partial \mathcal{L} (\{\mathbf{W}_i\}, \{\mathbf{b}_i\}) + G_f(f_A^*)}{\partial \mathbf{W}_i} + \lambda_i({\bf{W}}_{i}-{\bf{Z}}_{i} +\mathbf{U}_i)\\
\mathbf{b}_i^{k+1} &= \frac{\partial \mathcal{L} (\{\mathbf{W}_i\}, \{\mathbf{b}_i\}) + G_f(f_A^*)}{\partial \mathbf{b}_i}
\end{aligned}
\end{equation}
Since the loss of the classifier $\mathcal{L} (\{\mathbf{W}_i\}, \{\mathbf{b}_i\}$ is differentiable and $G_f(f_A^*)$ is also differentiable with respect to its input $f(\{\mathbf{W}_i\},\{\mathbf{b}_i\};x)$, the above equation has analytical solution.\\
The sub-problem of optimization $\bf{Z_i}$ can be written as
\begin{equation}
\begin{aligned}
& \underset{ \{{\bf{Z}}_{i}\}}{\text{min}} \quad
  \sum_{i=1}^{N} g_i(\mathbf{Z}_i)
 + \sum_{i=1}^{N} \frac{\lambda_i}{2} \parallel {\bf{W}}_{i}-{\bf{Z}}_{i} +\mathbf{U}_i \parallel_F^2
\end{aligned}
\end{equation}
where $g_i(\cdot)$ is the indicator function of whether the number of non-zero $\mathbf{Z}_i$ smaller than $n_i$. Intuitively, the optimal $\{\mathbf{Z}_i\}$ is a sparse approximation of $\mathbf{W}_i + \mathbf{U}_i$. According to \cite{zhang2018systematic}, the optimal solution for the above equation can be obtained by keeping the largest $n_i$ element of $\mathbf{W}_i^{k+1} + \mathbf{U}_i^{k}$ and set the rest of small values to zero.\\

\section{SECTION 4: Classification Model}
When training the classification model, the batch size is set as 64, and the training epoch is 300. The optimizer is Adam, and the learning rate is 0.05. For the MIA-Pruning, we pre-train the classification model for 200 epochs and then we follow the MIA-Pruning process shown in Algorithm 1 in the main text.

\section{SECTION 5: Inference Attack Model}
The inference attack model is composed of three fully-connected sub neural networks in a hierarchical structure. The prediction vector $f(x)$ and the targeted label $y$ are fed into two sub-networks in the first level in parallel, and the processed representations of the two sub-networks are then concatenated and fed into the third sub-network on the second level to make the final prediction. 
The architectures of the two sub-networks for processing $f(x)$ and $y$ are $[100, 1024, 512, 64]$ and $[100, 512, 64]$, respectively. The architecture of the third sub-network is $[256, 64, 1]$. We use ReLU as the activation function for the whole network. The weights are initialized following $\mathcal{N}(0, 0.01)$. We use Adam optimizer with the learning rate $0.001$. For CIFAR-10-CNN in Table 2 in the main text, the inference attack model consists of one fully-connected layer, the same as in \cite{rahman2018membership}.  For the training of the inference model in the Min-Max game (line 3-6 in Algorithm 1 in the main text), we use half of the training set and half of the testing set of the classification model as $D$ and $D'$. The batch size is 64, and we set $iterations$ to make it goes through roughly one epoch of the training data for the inference model for each epoch in line 2 in Algorithm 1 in the main text. We train the inference model using Adam optimizer with the learning rate as 0.00001. After training of the final classification model, we retrain the inference model for 100 epochs using another $D$ and $D'$ sampled differently from the classification model's training and validation set. 

\section{SECTION 6: Pruning Rate in Experiment}
In our experiment, we rank the weights based on the magnitude and prune a certain percentage of the smallest weights per layer by the following rules: 1) in the first layer, we prune 30\%-40\% 2) for the rest of the layers, we increase the percentage value from 50\%-95\%, depend on the depths and the structures of different DNN models. For example, on LeNet-5 model, the pruning rate for each layer is 40\%, 90\%, 93.75\%, 93.75\%, 93.75\%. We also show the total prune amount and rate in table 4 in the main text.
\section{SECTION 7: Extra Experiment results}
\begin{figure}[h]
\centering
	\includegraphics[width=0.3\textwidth]{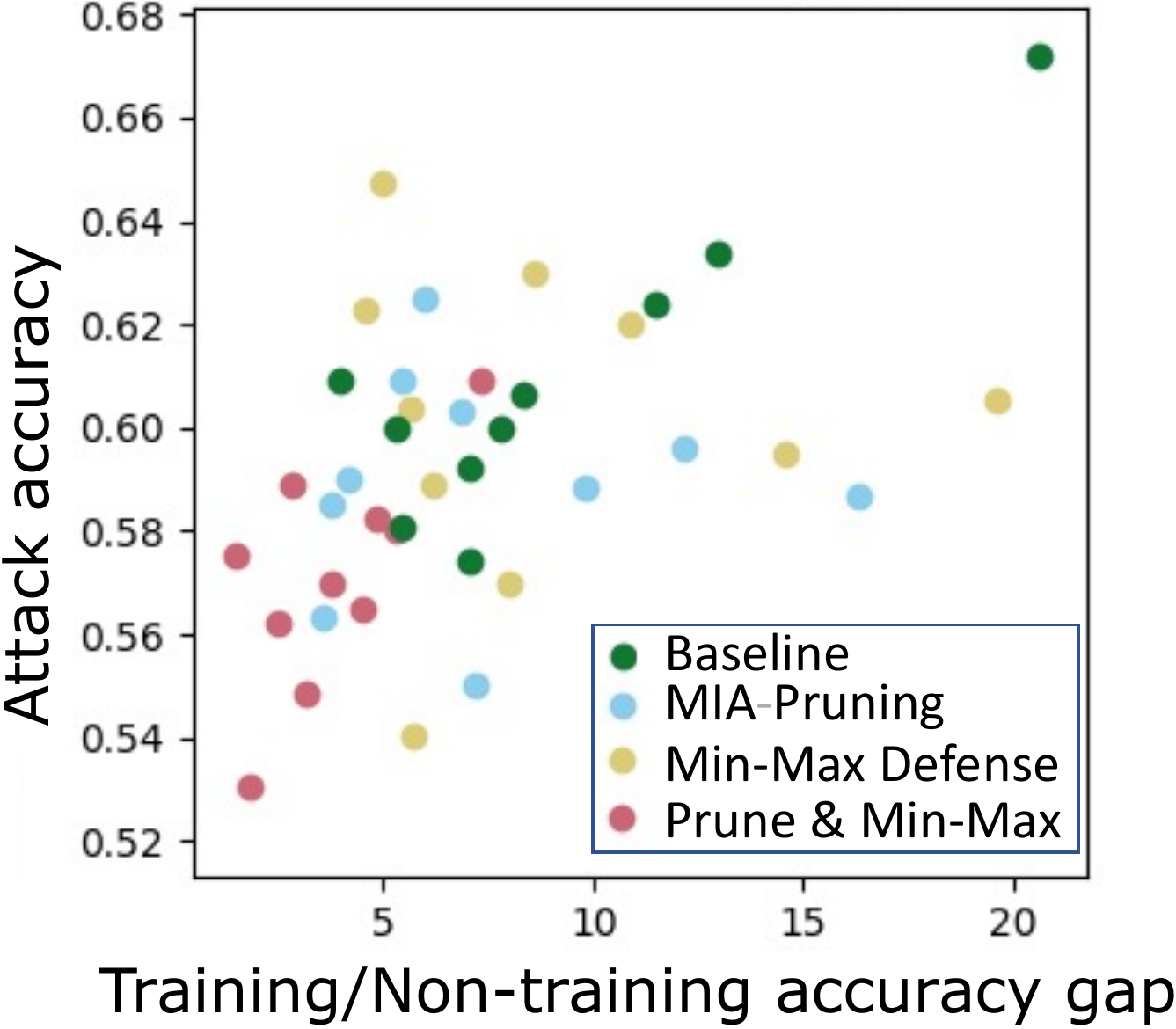}
    \caption{Attack accuracy versus training/non-training accuracy gap of VGG16 on CIFAR-10, which is defined as the difference of classification accuracy between training and non-training data.  }
    \label{fig:train_test_gap}
\end{figure}

To summarize the difference of prediction between training and non-training data quantitatively, we plot the MIA accuracy along with the difference of classification accuracy between training and non-training data for each class in CIFAR-10 in Figure \ref{fig:train_test_gap}. We name such difference as training/non-training accuracy gap. As shown in Figure \ref{fig:train_test_gap}, there is a correlation between the train/non-training accuracy gap and the attack accuracy. The larger the training-non-training accuracy gap is, the higher the membership attack accuracy is. Among all the four methods, MIA-Pruning \& Min-Max achieves the smallest train-test accuracy gap and lowest membership inference attack accuracy, therefore providing the highest privacy enhancement. 
%The comparison of overall training accuracy, testing accuracy, and membership inference attack accuracy is illustrated in Table \ref{tab:train_test_gap}, which conveys similar messages as Figure \ref{fig:train_test_gap}. 
\end{document}